\documentclass[11pt,a4paper]{article}
\usepackage[margin=2.2cm]{geometry}
\usepackage{amsmath,amssymb,bm,mathtools}
\usepackage{graphicx}
\usepackage{booktabs}
\usepackage{tabularx}
\newcolumntype{Y}{>{\raggedright\arraybackslash}X}
\usepackage{multirow}
\usepackage{array}
\usepackage{float}
\usepackage{caption}
\usepackage{placeins}
\usepackage{hyperref}
\hypersetup{hidelinks}
\usepackage{cite}
\usepackage{xcolor}
\usepackage[normalem]{ulem}

\newcommand{\diff}{\mathop{}\!\mathrm{d}}
\newcommand{\RMS}{\operatorname{RMS}}

\graphicspath{{figures/main/pdf/}{figures/main/png/}{figures/si/pdf/}{figures/si/png/}{../figures/main/pdf/}{../figures/main/png/}{../figures/si/pdf/}{../figures/si/png/}}
\title{AI4Plasma I: Data-Efficient Surrogate Modeling of Capacitively Coupled Plasma Fields}
\author{%
Zhenyu Zou$^{1,2}$, Junjue Liao$^{3}$, Hao Yang$^{1}$, Luowei Yin$^{1,3}$, Chenyu Wang$^{1}$,\\
Yufeng Wei$^{1,*}$, Chun Chen$^{2}$, Daniel~Zhengyu~Huang$^{4}$, Kangxin Liu$^{1}$,\\
Haoran Guan$^{1}$, Rubing Han$^{1}$, Yipei Chen$^{1}$, Peiting You$^{1}$, Yuwei Fan$^{1}$\\
[0.8em]
\small $^{1}$Theory Lab, Huawei Leibniz Research Center,
Shatin, N.T., Hong Kong SAR, China.\\
\small $^{2}$Department of Mechanical and Automation Engineering,
The Chinese University of Hong Kong,\\
\small Shatin, N.T., Hong Kong SAR, China.\\
\small $^{3}$Department of Mathematics,
The Hong Kong University of Science and Technology,\\
\small Clear Water Bay, Kowloon, Hong Kong SAR, China.\\
\small $^{4}$Beijing International Center for Mathematical Research,
Peking University, Beijing 100871, China.\\
[0.4em]
\small $^{*}$Corresponding authors: Yufeng Wei
(\href{mailto:wei.yufeng@huawei.com}{wei.yufeng@huawei.com}))
}
\date{}

\begin{document}

\maketitle

\begin{abstract}
As semiconductor devices move toward smaller feature sizes and high-aspect-ratio structures, plasma processing increasingly requires spatially resolved predictions of capacitively coupled plasma (CCP) states across operating conditions. Reference CCP simulations provide these fields, but generating converged solutions across a process window is costly. Surrogate models trained on simulation data offer inexpensive predictions, motivating comparison of reference-data requirements across architectures for prescribed accuracy. We systematically compare five surrogate models' data efficiency and examine how selected architectural components affect predictive accuracy. The models represent three approaches: coordinate-based pointwise prediction using a coordinate multilayer perceptron (Coordinate MLP), global modal reconstruction using a principal component analysis network (PCA-Net), and spatially coupled learning using U-Net, the Fourier neural operator (FNO), and Transolver. They predict cycle-averaged electron and ion densities, electric potential, and electron temperature for one-dimensional (1D) planar and two-dimensional (2D) axisymmetric Gaseous Electronics Conference (GEC) CCP benchmarks. FNO requires the fewest or joint-fewest cases in seven of eight comparisons at 1\% and 2\% four-field mean physical-space relative L2 error targets. For 1\% 1D interior error, FNO needs 50 cases versus Transolver's 100 and U-Net's 200. For 1\% 2D interior error, FNO and U-Net each require 100 cases. For 2\% 2D parameter-boundary error, FNO and Transolver require 200 cases versus U-Net's 400, representing an estimated 760 CPU core-hours of accepted simulations. For 1\% 2D boundary error, Transolver requires 400 cases versus FNO's 800. Structural controls assess FNO spectral modes and padding, U-Net skip connections, and Transolver adaptive slicing. PCA-Net prediction errors exceed matched rank-96 basis projection errors by factors of 9.78–13.53, showing that low-rank representation alone does not determine predictive accuracy. These results support FNO as the preferred surrogate for the studied CCP benchmarks when reference simulations are limited, while identifying specific high-accuracy objectives for which U-Net and Transolver provide stronger alternatives.
\end{abstract}


% ======================================================================
\section{Introduction}

As semiconductor devices move toward smaller feature sizes, higher aspect ratios, and more complex three-dimensional structures, plasma etching and deposition face increasingly stringent requirements for process-window control, uniformity, and the transport of charged particles and reactive species~\cite{ref1,ref2,ref45}. Operating conditions such as gas pressure, radio-frequency (RF) power or voltage, driving frequency, and gas composition affect etch rate, selectivity, uniformity, and feature profiles by modulating the plasma state~\cite{ref1,ref2,ref46}. Capacitively coupled plasma (CCP) is a widely used low-temperature plasma source in semiconductor manufacturing~\cite{ref2,ref45,ref48}. In CCPs, changes in operating conditions simultaneously affect sheath dynamics, electron heating, ionization, charged-particle transport, and plasma-wall interactions~\cite{ref5,ref6,ref4,ref48}, and consequently alter particle energies, fluxes, and spatial plasma distributions~\cite{ref7,ref8,ref9,ref48}. The mapping from operating conditions to plasma response is therefore inherently coupled and spatially resolved, and its reliable prediction is important for subsequent process analysis and operating-condition selection.

Physics-based forward models have long been developed to describe low-temperature plasma responses under prescribed operating conditions~\cite{ref6,ref49}. Early one- and two-dimensional Gaseous Electronics Conference (GEC) reference-cell studies, together with fluid, kinetic particle-in-cell (PIC), and hybrid simulations, established descriptions of bulk--sheath structure, charged-particle transport, and RF discharge dynamics~\cite{ref6,ref49,ref50,ref51,ref52,ref53}. Hybrid plasma equipment models developed by Kushner and co-workers further incorporated electron kinetics, heavy-species transport, electromagnetic fields, and surface reactions, linking chamber-scale plasma behavior with wafer-facing fluxes and feature-profile evolution~\cite{ref52,ref14,ref54,ref55,ref56}.

Multidimensional reactor models and numerical methods,
including parallel computing, graphics processing unit
(GPU) acceleration, implicit integration, and steady-state
acceleration, have expanded the spatial scales and problem
sizes that can be treated~\cite{ref50,ref54,ref57,ref58,ref59,ref60,ref53,ref61,ref62,ref63}.
Nevertheless, parameter sweeps, optimization, and inverse
design require simulations across many operating conditions,
while RF CCP models must generally be advanced to a periodic
steady state. The resulting computational cost remains
substantial. For example, Li \textit{et al.} discussed
two-dimensional CCP fluid simulations requiring hours to
days to reach convergence~\cite{ref63}, while Tsagkaridis
\textit{et al.} reported simulation times ranging from
several hours to days for an argon CCP model with detailed
collisional--radiative kinetics~\cite{refTsagkaridis2025}. In the present 2D CCP benchmark,
each accepted reference simulation requires an average of
3.80~CPU core-hours. Generating 200 and 800 reference cases
therefore corresponds to approximately 760 and 3042
CPU core-hours, respectively, excluding unsuccessful
attempts and retries. These costs highlight the importance
of reducing the number of reference simulations needed
to achieve a target prediction accuracy.

This computational burden has motivated the use of artificial intelligence in semiconductor process development. Work from Lam Research has demonstrated that human--machine collaboration can reduce the
cost-to-target in virtual plasma process development~\cite{ref35}. Other studies have combined plasma equipment simulations with surrogate models and optimization. Work from Samsung has used plasma simulations, deep-learning surrogates, multi-objective optimization, and simulation-guided digital twins for plasma process development~\cite{ref40,ref28}, while evolutionary optimization has also been used to search plasma operating conditions directly~\cite{ref66}. In simulation-based workflows, surrogate models can replace repeated forward calculations during parameter searches with low-cost predictions, thereby reducing the online computational cost~\cite{ref64}.

This acceleration, however, shifts a substantial part of the computational burden to reference-data generation. Each additional independent training sample typically requires another physics-based simulation. Previous reviews have highlighted the challenges of limited training data in plasma surrogate modeling~\cite{ref17,ref25,ref29}. Individual studies have further examined how predictive performance varies with training-data volume or coverage~\cite{ref31,ref27,ref18,ref67,ref33}. The practical value of a surrogate therefore depends not only on inference speed or the final error at a fixed training size, but also on the number of reference simulations required to reach a target accuracy. We refer to this relation between training-set size and prediction accuracy as data efficiency~\cite{ref67,ref32,ref33,ref41,refCost}. For multidimensional plasma simulations, generating hundreds of converged reference cases can itself constitute a substantial offline computational cost. Although active learning, physics-informed learning, multifidelity learning, and unsupervised pretraining have been developed to reduce expensive-data requirements~\cite{ref68,ref69,ref70,ref71,ref72}, different surrogate architectures can exhibit different error reductions as the training set grows~\cite{ref67,ref33,ref41}. For a prescribed accuracy target, the practical question is therefore
which architecture requires the fewest reference simulations and
whether that choice remains effective across CCP modeling settings
and operating-parameter regions.
Many existing comparisons emphasize performance at only one or
a few training-set sizes, leaving less clear how reference-data
requirements and model preferences vary across CCP benchmarks
and between parameter-interior and parameter-boundary predictions.

We systematically compare five representative surrogate architectures to determine
how many reference CCP simulations each requires to achieve
prescribed field-prediction accuracy, and how model
preference shifts across CCP modeling settings, reference-data
budgets, and interior versus parameter-boundary operating regimes.

The five architectures represent three approaches to spatial
field prediction: coordinate-based pointwise prediction using a
coordinate multilayer perceptron (Coordinate MLP)~\cite{ref38};
global modal reconstruction using a principal component analysis
network (PCA-Net)~\cite{ref36,ref19,bhattacharya2021model};
and spatially coupled learning using U-Net~\cite{ref23},
the Fourier neural operator (FNO)~\cite{ref22,ref21},
and Transolver~\cite{ref24}.
This selection allows us to examine how different representations
of spatial structure affect data efficiency.
Beyond cross-model comparisons, we perform targeted structural
controls to
identify the critical architectural mechanisms governing accuracy
and efficiency:
FNO spectral modes and padding, U-Net skip connections, and
Transolver adaptive slicing, while comparison with PCA projection
errors distinguishes low-rank representation accuracy from
prediction accuracy at unseen operating conditions.

We evaluate these architectures on complementary 1D planar and
2D axisymmetric GEC CCP benchmarks~\cite{ref49,ref50,ref51,ref57},
predicting RF-cycle-averaged electron density, ion density,
electric potential, and electron temperature.
The 1D planar model resolves electrode-normal bulk--sheath dynamics
and provides a computationally efficient setting for examining
operating-condition trends.
The 2D axisymmetric model additionally resolves radial
nonuniformity, geometric transitions, and chamber-scale field
distributions~\cite{ref50,ref57,ref15}.
Together, these physically and numerically distinct benchmarks
allow us to assess whether surrogate preferences differ
between the two CCP modeling settings.

To connect surrogate performance to the underlying plasma
physics, we first characterize the reference CCP fields
through their spatial distributions, operating-condition
responses, and PCA spectra. We then compare learning curves
and reference-data requirements at the 1\% and 2\% error
targets, examine the spatial distribution of prediction errors,
and investigate representation-specific behavior through
structural controls and PCA-based reconstruction analysis.
Finally, we discuss practical criteria for selecting surrogate
models according to reference-simulation budgets and
prediction objectives.

\section{Methods}

\subsection{Reference CCP simulations and data generation}

\paragraph{Simulation setup.}
Two reference CCP benchmarks are considered: a 1D planar CCP and a 2D
axisymmetric GEC CCP~\cite{ref49,ref50}. Both describe Ar discharges over
RF voltages of $U_{\mathrm{RF}}=100$--$200~\mathrm{V}$, gas pressures of
$P=0.5$--$2~\mathrm{Torr}$, and driving frequencies of
$f=6.78$--$54.24~\mathrm{MHz}$. The 1D model uses a parallel-plate geometry
with a $25.4~\mathrm{mm}$ electrode gap and 128 cells, while the 2D model is
stored on a $130\times40$ ($z\times r$) grid with 4624 active plasma cells.

\begin{figure}[!htbp]
\centering
\includegraphics[width=0.90\textwidth]{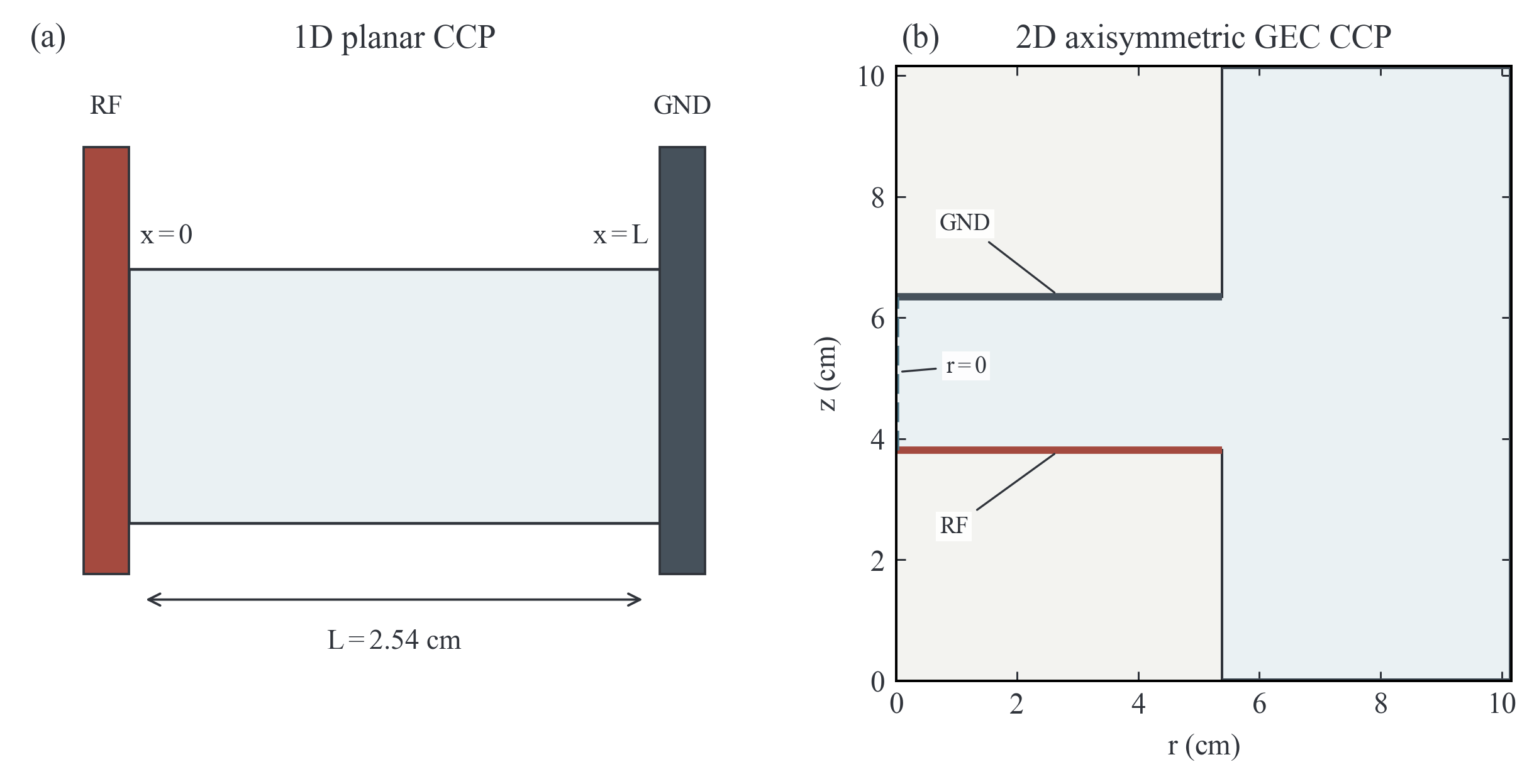}
\caption{Geometries and principal boundary conditions of the reference CCP
benchmarks: (a) 1D planar CCP and (b) 2D axisymmetric GEC CCP.}
\label{fig:cases}
\end{figure}

\paragraph{Governing equations.}
The reference simulations use a fluid description that self-consistently
solves the charged-particle continuity equations, electron energy equation,
and Poisson equation~\cite{ref6,ref49,ref50}:
\begin{equation}
\label{eq:compact_plasma_model}
\begin{aligned}
\frac{\partial n_s}{\partial t}
+\nabla\cdot\boldsymbol{\Gamma}_s &=S_s,
&&s\in\{e,i\},\\
\frac{\partial \varepsilon_e}{\partial t}
+\nabla\cdot\boldsymbol{\Gamma}_{\varepsilon} &=Q_e,\\
\boldsymbol{\Gamma}_e&=-\mu_e n_e\boldsymbol{E}-D_e\nabla n_e,\\
\boldsymbol{\Gamma}_i&=\mu_i n_i\boldsymbol{E}-D_i\nabla n_i,\\
-\nabla\cdot(\varepsilon_0\nabla\phi)&=e(n_i-n_e),
&&\boldsymbol{E}=-\nabla\phi.
\end{aligned}
\end{equation}
Here $s\in\{e,i\}$ labels electron or ion species, $n_s$ is number density,
$\boldsymbol{\Gamma}_s$ is particle flux, $S_s$ is the corresponding source,
$\varepsilon_e$ and $\boldsymbol{\Gamma}_{\varepsilon}$ are electron energy
density and energy flux, and $Q_e$ is the energy source. The coefficients
$\mu_s$ and $D_s$ denote mobility and diffusion, $\varepsilon_0$ is vacuum
permittivity, and $e$ is the elementary charge. The electric field
$\boldsymbol{E}$ is obtained from the electric potential $\phi$.

For the electron temperature $T_e$ expressed numerically in electronvolts (eV),
the electron energy density is $\varepsilon_e=3n_e eT_e/2$ in SI units;
if $T_e$ is expressed in kelvin, the equivalent relation is
$\varepsilon_e=3n_e k_{\mathrm B}T_e/2$, where $k_{\mathrm B}$ is the
Boltzmann constant. The background Ar temperature is fixed at
$T_g=300~\mathrm{K}$, and the neutral density follows
$n_{\mathrm{Ar}}=P/(k_{\mathrm B}T_g)$.
Electron-impact ionization and excitation, charged-particle
wall losses, and secondary electron emission are included. Further transport,
reaction, wall-model, convergence, and numerical-verification details are
provided in Appendix~\ref{app:verification}.

\paragraph{Quantity of interest.}
For each operating point $(U_{\mathrm{RF}},P,f)$, we denote the spatial
coordinate by $\boldsymbol{x}$, with its single component $x$ in 1D and
$\boldsymbol{x}=(z,r)^{\mathsf T}$ in 2D. Let $\Omega$ denote the plasma
domain. The instantaneous field vector is
\[
\boldsymbol{y}_{\mathrm{inst}}(U_{\mathrm{RF}},P,f;\boldsymbol{x},t)
=\bigl(n_e,n_i,\phi,T_e\bigr)^{\mathsf T}.
\]
Here $n_e$ and $n_i$ are the electron and ion number densities,
$\phi$ is the electric potential, and $T_e$ is the electron temperature.
The semicolon separates the operating conditions from the spatial coordinates.

The reference simulations are advanced toward periodic steady state.
Convergence is assessed using six diagnostics: the spatial means of $n_e$,
$n_i$, and $T_e$, the maxima of $n_e$ and $n_i$, and the spatial integral
of $n_i$. The 2D solver additionally monitors the maximum $T_e$ and the
relative $L_2$ change of its spatial profile. Scalar diagnostics and,
in 2D, cycle-averaged $T_e$ profiles are averaged over consecutive,
non-overlapping blocks of 20 RF periods. Starting at 300 RF periods,
the largest relative change between adjacent blocks must remain below
$5\times10^{-3}$ for three consecutive checks. The supervised
targets are obtained by averaging over the final complete RF period,
\begin{equation}
\boldsymbol{y}(U_{\mathrm{RF}},P,f;\boldsymbol{x})
=
\frac{1}{T_{\mathrm{RF}}}
\sum_{n\in\mathcal{C}_{\mathrm{last}}}
\boldsymbol{y}_{\mathrm{inst}}^{(n)}(U_{\mathrm{RF}},P,f;\boldsymbol{x})\Delta t_n .
\label{eq:cycle_average_new}
\end{equation}
Here $T_{\mathrm{RF}}$ is the RF period, $\mathcal{C}_{\mathrm{last}}$
is the set of time-step indices in the final complete period,
$\boldsymbol{y}_{\mathrm{inst}}^{(n)}$ is the instantaneous field at step $n$, and
$\Delta t_n$ is the time-step duration. Thus $\boldsymbol y$ always denotes
the cycle-averaged fields used as learning targets.

\paragraph{Training dataset.}
Operating points are generated by centered Latin hypercube sampling in the
$U_{\mathrm{RF}}$--$P$--$f$ space. The resulting datasets contain 3000 valid 1D cases
(2400 training-pool, 300 interior-test, and 300 parameter-boundary cases) and
1000 valid 2D cases (800, 100, and 100, respectively).

To distinguish interior from parameter-boundary prediction, $U_{\mathrm{RF}}$, $P$, and $f$
are normalized to $[0,1]$ (denoted by tildes), and the distance to the nearest edge of the
parameter domain is defined as
\[
d_{\mathrm{bdry}}
=\min\left\{\widetilde{U}_{\mathrm{RF}},1-\widetilde{U}_{\mathrm{RF}},
\tilde P,1-\tilde P,\tilde f,1-\tilde f\right\}.
\]
This dimensionless quantity measures the smallest normalized distance to any of the six faces of the rectangular $U_{\mathrm{RF}}$--$P$--$f$ range. The parameter-boundary set contains the cases with the smallest
$d_{\mathrm{bdry}}$, whereas interior cases satisfy
$d_{\mathrm{bdry}}\geq0.10$. Both sets remain within the prescribed operating
ranges; the parameter-boundary evaluation therefore probes prediction near the
edge of the process window rather than extrapolation beyond it. The two test
sets are fixed before model comparison and are identical for all models.

In our dataset, the operating parameters and model-specific spatial coordinates are
standardized using statistics from the current training subset. In 1D,
$n_e$ and $n_i$ are transformed using $\log_{10}$; in 2D, the same
transformation is additionally applied to $T_e$, while the remaining outputs
remain in linear form. All normalization statistics and PCA bases are obtained
exclusively from the corresponding training subset.

\paragraph{Evaluation protocol.}
Data efficiency is evaluated at the following reference training-set sizes:
\[
\begin{aligned}
\mathcal{N}_{\mathrm{1D}}&=\{50,100,200,400,800,1600,2400\},\\
\mathcal{N}_{\mathrm{2D}}&=\{25,50,100,200,400,600,800\}.
\end{aligned}
\]
Each $N\in\mathcal{N}_{\mathrm{1D}}$ or $N\in\mathcal{N}_{\mathrm{2D}}$ denotes the number of reference training cases. For each geometry, training-set size, and repeat, all five models use the same training
cases and fixed test sets. In 1D, training-case identities are fixed across
repeats and only the optimization seed varies. In 2D, initialization and
data-order seeds vary, and for $N<800$ the selected training subset also varies
across repeats; within each repeat and training-set size, however, all models share the
same subset. At $N=800$, the full 800-case training pool is used.

The reference datasets were generated on nodes with two 32-core 7285H
processors at 2.5~GHz and 256~GB DDR4 memory. Each case was run serially on one
central processing unit (CPU) core. Table~\ref{tab:data_protocol} summarizes the benchmark setup and
reference-simulation cost. The mean successful-attempt wall time over the accepted reference cases is 1.7197~min per 1D case and 3.8024~h per 2D case. These exclude failed attempts and retry overhead. Thus $N$ times this mean provides an accounting estimate of reference-generation cost, rather than the exact time of a selected subset.

\begin{table}[!htbp]
\centering\small
\caption{Reference-simulation setup, data sizes and computational cost for the two CCP benchmarks.}
\label{tab:data_protocol}
\setlength{\tabcolsep}{5pt}
\begin{tabularx}{\linewidth}{@{}p{0.31\linewidth}XX@{}}
\toprule
Item & 1D planar CCP & 2D axisymmetric GEC CCP\\
\midrule
Geometry & Parallel plates, $L=25.4~\mathrm{mm}$ & Axisymmetric GEC chamber\\
Operating conditions & $U_{\mathrm{RF}}=100$--$200~\mathrm{V}$; $P=0.5$--$2~\mathrm{Torr}$; $f=6.78$--$54.24~\mathrm{MHz}$ & Same parameter ranges\\
Spatial grid & 128 cells & $130\times40$ ($z\times r$); 4624 active cells\\
Target fields & $n_e,n_i,\phi,T_e$ (RF-cycle averaged) & $n_e,n_i,\phi,T_e$ (RF-cycle averaged)\\
Valid reference cases & 3000 & 1000\\
Train / interior / boundary & $2400/300/300$ & $800/100/100$\\
Tested training-set sizes & 7, $N=50$--2400 & 7, $N=25$--800\\
Mean accepted-attempt CPU wall time & $1.7197~\mathrm{min/case}$ & $3.8024~\mathrm{h/case}$\\
\bottomrule
\end{tabularx}
\end{table}

All predictions are transformed back to physical units before evaluation.
For test case $c\in\mathcal C$, field $k\in\{1,\ldots,4\}$,
and repeat $q\in\{1,\ldots,5\}$, let $\boldsymbol u_{c,k}$ and
$\hat{\boldsymbol u}_{c,k}^{(q)}$ be the reference and predicted vectors
of cycle-averaged values at the active cells. The relative $L_2$ error is
\begin{equation}
e_{c,k}^{(q)}
=
\frac{
\left\|\hat{\boldsymbol u}_{c,k}^{(q)}-\boldsymbol u_{c,k}\right\|_2
}{
\max\!\left(\left\|\boldsymbol u_{c,k}\right\|_2,\varepsilon_k\right)
}.
\label{eq:field_error}
\end{equation}
The active-cell vectors have 128 components in 1D and 4624 components in
2D; $\|\cdot\|_2$ denotes the Euclidean norm. The positive numerical safeguard $\varepsilon_k$ has the same physical
units as $\|\boldsymbol u_{c,k}\|_2$; in the reported unit system its
stored numerical value is $10^{-30}$ for each field. The four fields and
all test cases are weighted equally,
\begin{equation}
e_N^{(q)}
=
\frac{1}{4|\mathcal C|}
\sum_{c\in\mathcal C}\sum_{k=1}^{4}e_{c,k}^{(q)}.
\label{eq:mean_four_field_error}
\end{equation}
For each model and training-set size $N$, five independent trainings are
evaluated. We report the median and interquartile range (IQR, from the first
to the third quartile) of $e_N^{(q)}$ across repeats. Data efficiency is assessed
by how rapidly this error decreases as the number of reference cases increases.

To quantify the amount of reference data required to reach a common accuracy,
we define
\begin{equation}
N_{\tau}=\min\left\{N\in\mathcal N:
\operatorname{median}_{q=1,\ldots,5}\!\left(e_N^{(q)}\right)\leq\tau\right\},
\qquad \tau\in\{0.01,0.02\},
\label{eq:n1percent}
\end{equation}
where $\mathcal N$ is the set of tested training-set sizes for the
corresponding benchmark. The minimum is defined only if at least one tested
training-set size meets the target; otherwise we report NR (not reached).
The symbols $N_{1\%}$ and $N_{2\%}$ correspond to $\tau=0.01$ and
$\tau=0.02$, respectively. No interpolation between tested training-set sizes is used. The percentages
are convenient comparison targets, not manufacturing acceptance criteria. The reported errors quantify agreement with the fixed discretized CCP
reference fields; their numerical resolution dependence is reported separately
in Appendix Table~\ref{tab:numerical_sensitivity}.

\subsection{Spatial representations in surrogate models}

\begin{figure}[!htbp]
\centering
\includegraphics[width=0.92\linewidth]{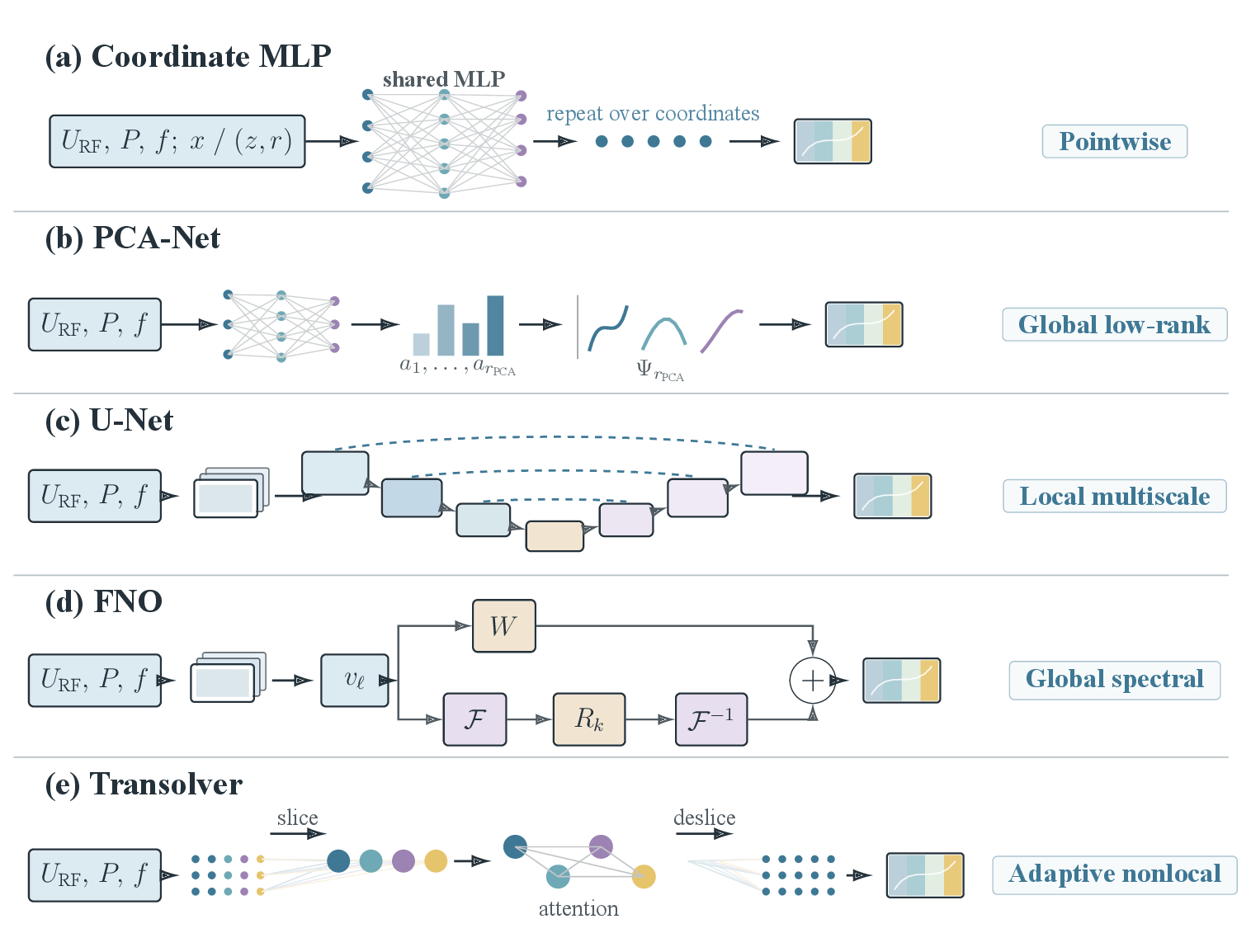}
\caption{Spatial representations of the five surrogate architectures.
Coordinate MLP predicts fields pointwise, PCA-Net reconstructs fields
using a global PCA basis, and U-Net, FNO, and Transolver exchange
spatial information through multiscale convolutions, Fourier modes,
and adaptive spatial grouping, respectively.}
\label{fig:architecture}
\end{figure}

Surrogate architectures differ in how they represent spatial
structure and exchange information across the CCP domain.
To examine how these differences relate to data efficiency,
we consider five architectures spanning three complementary
strategies. For operating conditions $(U_{\mathrm{RF}},P,f)$,
the target is the cycle-averaged field
$\boldsymbol{y}(U_{\mathrm{RF}},P,f;\boldsymbol{x})$
defined in Sec.~2.1, with $\hat{\boldsymbol y}$ denoting
the corresponding surrogate prediction in physical units. Coordinate MLP represents the fields as a function of
operating conditions and spatial coordinates, while PCA-Net
reconstructs them through a global low-rank basis.
U-Net, FNO, and Transolver learn spatially coupled feature
representations, but exchange spatial information differently:
through multiscale convolutions, Fourier modes, and adaptive
attention-based grouping, respectively. Together, these
architectures provide a representative comparison of
coordinate-based, reduced-order, and spatially coupled
approaches, which encode distinct spatial inductive biases.

Coordinate MLP directly approximates the field as a function of the operating
parameters and spatial coordinates. Writing the reconstructed prediction in
physical units,
\[
\hat{\boldsymbol y}(U_{\mathrm{RF}},P,f;\boldsymbol x)
=g_{\theta}(U_{\mathrm{RF}},P,f,\boldsymbol x).
\]
Here $\theta$ denotes the trainable network parameters; the expression
includes the inverse output transformation applied after network prediction,
as described in Sec.~2.3. The same network is evaluated at different
spatial locations to reconstruct the full field, without requiring a
predefined spatial basis or grid topology. Spatial dependencies are
thus learned implicitly through the shared coordinate function rather
than an explicit spatial information-exchange operation~\cite{ref38,ref74}.

PCA-Net instead represents the solution family in a global low-dimensional
subspace, combining a PCA basis with a learned coefficient
map~\cite{bhattacharya2021model}, following the broader ideas of
proper orthogonal decomposition and non-intrusive reduced-order
modeling~\cite{ref36,ref19,ref20},
\begin{equation}
\hat{\boldsymbol z}(U_{\mathrm{RF}},P,f)
=
\bar{\boldsymbol z}
+
\boldsymbol{\Psi}_{r_{\mathrm{eff}}}\boldsymbol a_{\theta}(U_{\mathrm{RF}},P,f),
\label{eq:pca_representation_story}
\end{equation}
where $\boldsymbol z\in\mathbb R^{4M}$ is the vector formed by
concatenating the four transformed and standardized reference fields
over $M$ active cells, and $\hat{\boldsymbol z}$ is its predicted
reconstruction in the same space. The vector
$\bar{\boldsymbol z}\in\mathbb R^{4M}$ is the training mean, and
$\boldsymbol\Psi_{r_{\mathrm{eff}}}\in\mathbb R^{4M\times r_{\mathrm{eff}}}$
contains the retained modes of the joint four-field PCA basis.
The network predicts their coefficients
$\boldsymbol a_{\theta}\in\mathbb R^{r_{\mathrm{eff}}}$.
Here $r_{\mathrm{eff}}$ is the effective retained rank for the current
training subset, as defined in Sec.~2.3. The reconstructed vector is
inverse-transformed to obtain the physical fields. The fixed PCA basis
imposes a low-rank constraint in the transformed field space: insufficient
modal coverage introduces a projection error, while reconstruction at
unseen operating conditions also depends on the predicted coefficients~\cite{ref42}.

We next consider the three architectures that explicitly couple spatial
features through different operations. U-Net uses local convolutions
together with multiscale encoder--decoder coupling,
\[
\boldsymbol e_{s+1}
=\mathcal D_s\!\left(\mathcal C_s(\boldsymbol e_s)\right),
\qquad
\boldsymbol d_s=\mathcal C'_s\!\left(
\left[\mathcal U_s(\boldsymbol d_{s+1}),\boldsymbol e_s\right]
\right).
\]
Here $\boldsymbol e_s$ and $\boldsymbol d_s$ are encoder and decoder
feature arrays at scale $s$, $\mathcal C_s$ and $\mathcal C'_s$ are convolutional blocks,
$\mathcal D_s$ and $\mathcal U_s$ are downsampling and upsampling operations, and square
brackets denote concatenation of feature channels. Local receptive fields are
progressively expanded across scales, while skip connections retain fine
spatial information. This introduces a bias toward local and hierarchical
spatial dependencies, giving U-Net a local multiscale representation~\cite{ref23,ref37,ref43,ref44}.

FNO introduces spatial coupling through Fourier modes,
\begin{equation}
\boldsymbol v_{\ell+1}(\boldsymbol x)
=
\sigma\!\left[
\boldsymbol W_{\ell}\boldsymbol v_{\ell}(\boldsymbol x)
+
\mathcal F^{-1}
\left(\boldsymbol R_{\ell}\mathcal F(\boldsymbol v_{\ell})\right)(\boldsymbol x)
\right].
\label{eq:fno_story}
\end{equation}
Here $\boldsymbol v_\ell$ is the feature vector at layer $\ell$,
$\boldsymbol W_\ell$ is a learned pointwise linear map, $\boldsymbol R_\ell$ acts on the retained
Fourier coefficients, $\mathcal F$ and $\mathcal F^{-1}$ denote the
Fourier transform and its inverse, and $\sigma$ is the activation function.
Because the retained Fourier modes have global support, spatially separated
regions can interact within each spectral layer. FNO therefore introduces
a global spectral bias, in contrast to the progressively enlarged local
receptive fields of U-Net~\cite{ref21,ref22}. Unlike PCA-Net, its predicted
fields are not restricted to a fixed low-dimensional subspace, since
pointwise transformations and nonlinear activations are also involved.
Fourier neural operators have also been applied to plasma surrogate
modeling, including limited-data settings~\cite{ref31,ref33}.
Geometric extensions such as Geo-FNO use learned coordinate deformations
to accommodate irregular physical domains~\cite{li2023geofno}.
The FNO considered here instead uses a fixed structured $z$--$r$
grid and an active-domain mask.

Transolver uses input-dependent slicing to aggregate spatial points into
learned spatial tokens, performs attention in the resulting token space, and maps
the updated information back to the physical domain~\cite{ref24},
\[
\boldsymbol T=\widetilde{\boldsymbol S}^{\mathsf T}\boldsymbol H,\qquad
\boldsymbol T'=\operatorname{Attn}(\boldsymbol T),\qquad
\boldsymbol H'=\boldsymbol S\boldsymbol T'.
\]
Here $\boldsymbol H\in\mathbb R^{M\times d}$ contains features at the $M$
active spatial points with $d$ feature channels, and
$\boldsymbol S\in\mathbb R^{M\times m}$ gives the weights of $m$ learned
slices. The slice weights $\boldsymbol S$ are nonnegative, and
$\widetilde S_{ij}=S_{ij}/\sum_{a=1}^{M}S_{aj}$ denotes the weights
normalized across points for each slice $j$, with $i$ indexing
active spatial points. The aggregation thus forms
weighted-average token features
$\boldsymbol T\in\mathbb R^{m\times d}$. Primes denote updated
features. The formula illustrates the slice--attention--deslice sequence;
residual connections and other implementation details are omitted.
Unlike fixed local neighborhoods or a fixed Fourier basis, this
representation organizes nonlocal spatial interactions through
input-dependent grouping.

We first characterize the spatial structure of the reference CCP
fields, then compare the data efficiency of the five surrogate
architectures and examine their representation-specific behavior.

\subsection{Training and evaluation}
\label{subsec:training_evaluation}

Within each CCP benchmark, the five models are compared at the same
training-set sizes and evaluated on the same held-out test sets,
with five training repeats at each size.

\paragraph{Training objectives.}
Coordinate MLP, U-Net, FNO, and Transolver are trained to predict the
transformed, standardized fields. Let $b\in\{1,\ldots,B\}$ index the cases
in a training batch, $k\in\{1,\ldots,4\}$ the predicted fields,
and $\Omega_b$ the training-cell locations selected for case $b$.
The targets $z_{b,k}$ and predictions $\hat z_{b,k}$ define the loss
\[
\mathcal L_{\mathrm{field}}=
\frac{\displaystyle\sum_{b=1}^{B}\sum_{k=1}^{4}
\sum_{\boldsymbol x\in\Omega_b}
[\hat z_{b,k}(\boldsymbol x)-z_{b,k}(\boldsymbol x)]^2}
{4\displaystyle\sum_{b=1}^{B}|\Omega_b|}.
\]
All 128 cells are used in 1D. In 2D, U-Net, FNO, and Transolver use
the 4624 active plasma cells, while Coordinate MLP samples 512 active
positions per case during training. All five models are evaluated on the
same complete active domain. The training loss uses transformed variables;
the benchmark errors in Eqs.~\eqref{eq:field_error} and
\eqref{eq:mean_four_field_error} are calculated after conversion back to
physical units.

PCA-Net instead predicts coefficients of the training-set PCA basis and
minimizes
\[
\mathcal L_{\mathrm{PCA}}=
\frac{1}{Br_{\mathrm{eff}}}
\sum_{b=1}^{B}
\|\hat{\boldsymbol a}_b-\boldsymbol a_b\|_2^2,
\qquad
r_{\mathrm{eff}}=\min\{r_{\mathrm{target}},N-1,r_{\mathrm{available}}\},
\]
where $\boldsymbol a_b$ and $\hat{\boldsymbol a}_b$ are the target and
predicted PCA coefficient vectors, $r_{\mathrm{eff}}$ is the number of
retained modes, $r_{\mathrm{target}}$ is the requested rank, $N$ is the
number of current training cases, and $r_{\mathrm{available}}$ is the
number of components returned by the fitted PCA decomposition.
The requested rank is 32 in 1D, giving $r_{\mathrm{eff}}=32$ for all tested
training sizes. In 2D it is 96, giving $r_{\mathrm{eff}}=24$, 49, and 96
for $N=25$, 50, and $N\ge100$, respectively.

\paragraph{Model selection and benchmark protocol.}
All networks are optimized with AdamW. To improve each model's predictive performance, hyperparameters and training schedules
are selected for each architecture and CCP benchmark using development or
validation data, without using either formal test set. After selection,
the adopted configuration is fixed across all reported training sizes and
repeats of that benchmark. The five models share the same training cases
within each benchmark, size, and repeat, as described in Sec.~2.1.
The comparison matches the number of expensive reference simulations,
not network parameter count, batch size, or GPU training time. Adopted
configurations and development records are given in
Appendix~\ref{app:data_methods}.

\paragraph{Structural sensitivity experiments.}
The 2D FNO benchmark uses $32\times20$ Fourier modes, with 10\% axial
and 5\% outer-radial padding. A separate mode-configuration study compares
$8\times5$, $16\times10$, $24\times15$, and $32\times20$ modes at $N=800$
with adjusted network widths. The FNO padding control uses $N=200$ and a
fixed 20-case development set, with two repeats and 1000 epochs. U-Net and
Transolver structural controls use this same 20-case development partition,
held out from the 800-case 2D training pool, with fitting sizes $N=25$,
200, and 780, two repeats, and 1000 epochs. U-Net is compared with no-skip
and single-scale convolutional neural network (CNN) variants; Transolver
is compared against its no-attention and fixed-radial-slicing variants. These experiments
examine structural sensitivity and do not use the formal test sets for
configuration selection.

\paragraph{Reference-field analyses.}
Principal component analysis (PCA) 
is performed on the reference training pools, with held-out reference fields
used to evaluate projection errors. For each physical field $u_k$, we
measure its variation across the training operating conditions
$\mathcal D$ by
\[
c_{\mathrm{var}}(\boldsymbol x)=
\frac{\operatorname{std}_{(U_{\mathrm{RF}},P,f)\in\mathcal D}
[u_k(U_{\mathrm{RF}},P,f;\boldsymbol x)]}
{\RMS_{(U_{\mathrm{RF}},P,f)\in\mathcal D,\,\boldsymbol x\in\Omega}
[u_k(U_{\mathrm{RF}},P,f;\boldsymbol x)]},
\]
where $\operatorname{std}$ is the population standard deviation across
operating conditions, $\RMS$ denotes root-mean-square, and the denominator
uses the training cases and active cells of the same benchmark. The spatially
averaged measure and the ratio between benchmarks are
\[
C_{\mathrm{var}}=\RMS_{\boldsymbol x\in\Omega}
[c_{\mathrm{var}}(\boldsymbol x)],\qquad
R_{\mathrm{var}}=\frac{C_{\mathrm{var}}^{\mathrm{2D}}}
{C_{\mathrm{var}}^{\mathrm{1D}}}.
\]
The field index $k$ is omitted from
$c_{\mathrm{var}}$, $C_{\mathrm{var}}$, and
$R_{\mathrm{var}}$ for brevity.
The PCA rank sweep fits field-wise bases on the corresponding training pool.
For a reference field $\boldsymbol u$ and its rank-$r_{\mathrm{PCA}}$
reconstruction $\boldsymbol u_{\mathrm{proj}}^{(r_{\mathrm{PCA}})}$,
we report the physical-space projection error
\begin{equation}
e_{\mathrm{proj}}^{(r_{\mathrm{PCA}})}=
\frac{\|\boldsymbol u-\boldsymbol u_{\mathrm{proj}}^{(r_{\mathrm{PCA}})}\|_2}
{\|\boldsymbol u\|_2}.
\label{eq:pca_projection_error}
\end{equation}
When a basis is fitted to transformed variables, the reconstruction is
inverse-transformed before evaluation; such a reconstruction is not
necessarily an orthogonal projection in physical units.

% ======================================================================
\section{Results and discussion}
\label{sec:results_new}

The central finding is that FNO requires the lowest or joint-lowest number of reference cases in seven of eight comparisons at the 1\% and 2\% error targets. More stringent accuracy targets, however, can substantially increase reference-simulation costs and favor different models. We first characterize the reference CCP fields through their spatial distributions, operating-condition responses, and PCA-based variability, and then compare the reference-data requirements of five surrogate models. 
Representation-specific analyses assess selected architectural
components of FNO, U-Net, and Transolver, while distinguishing
basis projection accuracy from predictive accuracy for PCA-Net. These results inform the model-selection recommendations in Sec.~\ref{sec:selection_synthesis}.

\subsection{Reference CCP characteristics}
\label{subsec:reference_ccp_characteristics}

\begin{figure}[!htbp]
\centering
\includegraphics[width=0.92\linewidth]{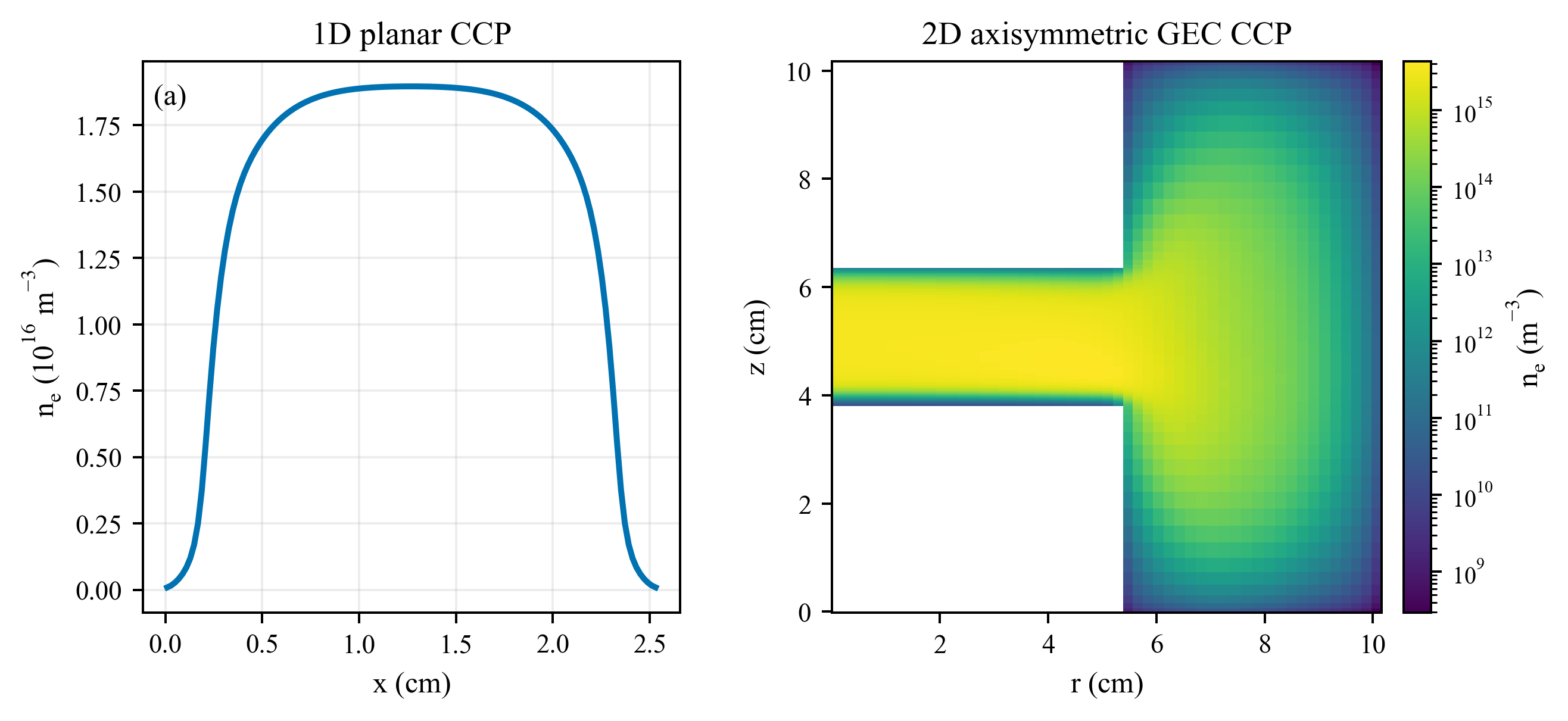}
\caption{Representative RF-cycle-averaged electron-density fields from
the reference CCP simulations: (a) 1D planar CCP and (b) 2D
axisymmetric GEC CCP. The 1D field exhibits predominantly axial
bulk--sheath variation, whereas the 2D field also shows radial
expansion and gradients associated with geometric transitions.}
\label{fig:reference_fields}
\end{figure}

The reference CCP fields show different spatial structures in the
1D and 2D benchmarks. The 1D fields exhibit axial bulk--sheath
variation, whereas the 2D GEC fields also exhibit radial variation
and gradients near geometric transitions. Figure~\ref{fig:reference_fields} shows
representative cycle-averaged electron-density fields for the two benchmarks.
In the 1D planar CCP, $n_e(x)$ forms a relatively flat high-density region in
the plasma bulk and decreases rapidly near the sheaths at both electrodes,
giving the characteristic bulk--sheath structure. Because only variations
normal to the electrodes are resolved, changes across operating conditions
mainly modify the bulk density, sheath extent, and axial gradients, while the
overall spatial organization remains similar. By contrast, the 2D
axisymmetric GEC CCP additionally exhibits radial expansion and
geometry-induced transitions. The high-density region expands from the narrow
channel into the wider chamber, producing pronounced gradients near the
channel exit, the expansion region, and the outer chamber. The 2D reference
solutions therefore contain radial and geometry-associated structures in
addition to the axial bulk--sheath organization resolved in 1D.

\begin{figure}[!htbp]
\centering
\includegraphics[width=0.75\linewidth]{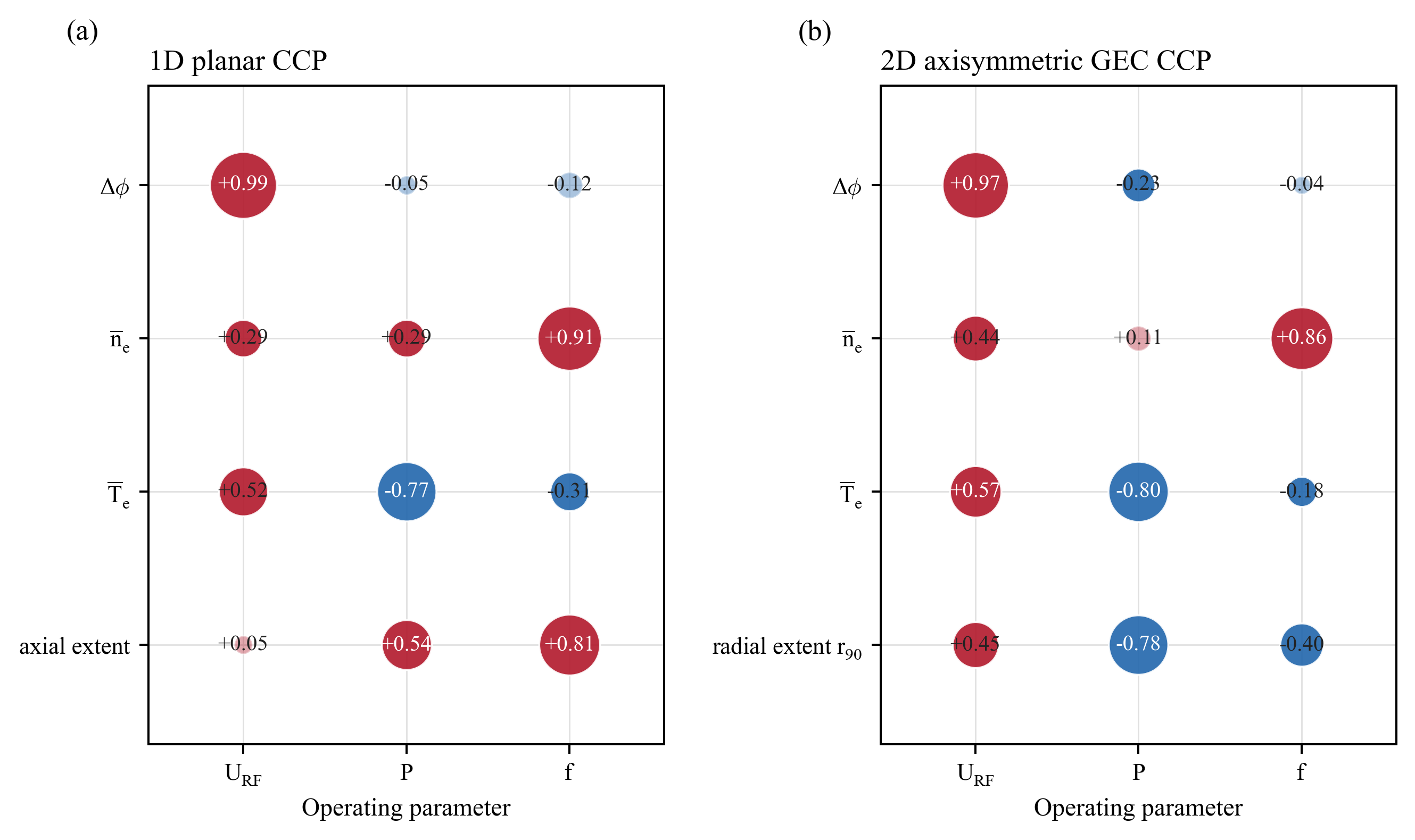}
\caption{Operating-condition dependence of the reference CCP fields.
Bubble matrices show signed Spearman correlations of the potential
range $\Delta\phi$, mean electron density $\overline{n}_e$,
mean electron temperature $\overline{T}_e$, and axial or radial
plasma-extent descriptors with RF voltage, pressure, and frequency.
Color indicates the sign and bubble area indicates $|\rho_s|$.
In both benchmarks, $\Delta\phi$ correlates strongly with RF voltage,
$\overline{n}_e$ with frequency, and $\overline{T}_e$ negatively
with pressure.}
\label{fig:parameter_response}
\end{figure}

The dependence of the reference CCP fields on operating conditions is examined
in Fig.~\ref{fig:parameter_response} using signed Spearman correlation
coefficients $\rho_s$ between $U_{\mathrm{RF}}$, $P$, and $f$ and selected plasma
descriptors. Here $\Delta\phi$ denotes the spatial potential range, while
$\overline{n}_e$ and $\overline{T}_e$ are arithmetic spatial means over the
corresponding domains (active cells in 2D). In both benchmarks, $\Delta\phi$
is strongly correlated with the applied RF voltage, with coefficients of
approximately $0.99$ in 1D and $0.97$ in 2D. The mean electron temperature
$\overline{T}_e$ is negatively correlated with pressure
($\rho_s\approx -0.77$ and $-0.80$), consistent with increased
electron--neutral collisions and inelastic energy losses at higher
pressures~\cite{ref48,ref5}. The mean electron density $\overline{n}_e$ is
positively correlated with driving frequency
($\rho_s\approx 0.91$ and $0.86$), consistent with higher cycle-averaged
plasma densities at higher frequencies over the operating range considered
here~\cite{ref48,ref9}.

Operating conditions also affect the spatial distribution of the plasma. In
1D, the axial extent is characterized by the coordinate containing 90\% of
the cumulative electron-density integral $n_e\,\diff x$. In 2D, $r_{90}$
denotes the smallest radius enclosing 90\% of the electron-density volume
integral $n_e\,\diff\mathcal V$, where
$\diff\mathcal V=2\pi r\,\diff r\,\diff z$; implementation details are given
in Appendix~\ref{app:data_methods}. The 1D axial extent is positively
correlated with frequency ($\rho_s\approx 0.81$), whereas the 2D radial
extent $r_{90}$ is negatively correlated with pressure
($\rho_s\approx -0.78$). Previous 2D CCP studies have also reported
radial nonuniformity and electrode-edge effects~\cite{ref4,ref15}.
The correlations observed here show that operating conditions affect both
the characteristic field levels and the spatial distributions that the
surrogate models must reconstruct.

\begin{figure}[!htbp]
\centering
\includegraphics[width=0.92\linewidth]{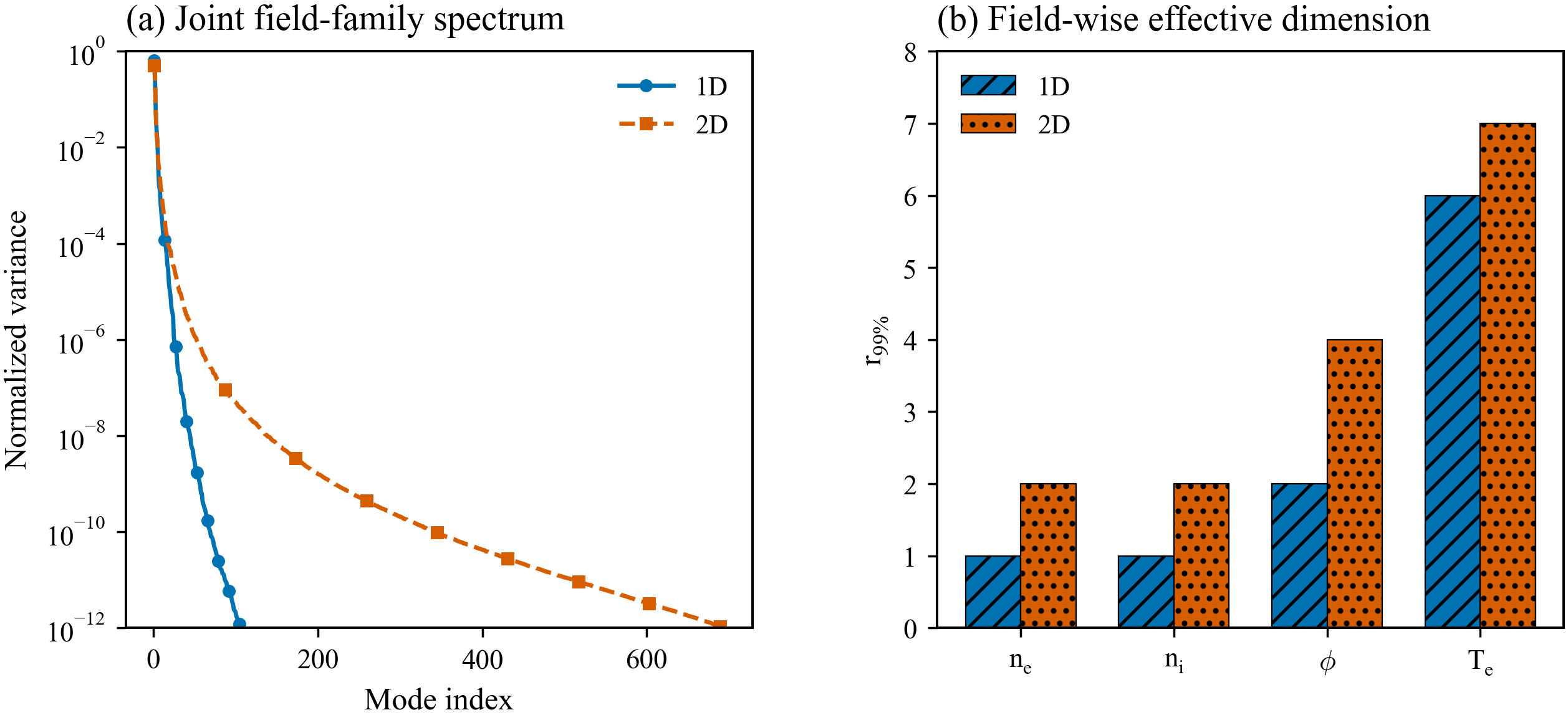}
\caption{PCA spectra of the reference CCP fields.
(a) Joint normalized explained-variance spectra for the
1D and 2D field families; (b) field-wise numbers of modes
$r_{99\%}$ required to capture 99\% of cumulative variance.
The 2D spectra decay more slowly and require more modes for
each field, although at most seven modes suffice to retain
99\% of variance in either benchmark.}
\label{fig:solution_manifold_final}
\end{figure}

The spatial organization of these reference fields can also be
characterized through PCA and their variability across operating conditions.
Figure~\ref{fig:solution_manifold_final} compares the spatial
variations of the 1D and 2D reference fields through PCA.
Overall, the explained-variance spectra decay more slowly
in 2D, and the number of modes required to capture 99\%
of the cumulative variance, $r_{99\%}$, increases for
all four fields. Specifically, $r_{99\%}$ increases from
1 to 2 for both $n_e$ and $n_i$, from 2 to 4 for
$\phi$, and from 6 to 7 for $T_e$. These differences
are consistent with the additional radial and
geometry-associated structures observed in
Fig.~\ref{fig:reference_fields}. Nevertheless, no more
than seven PCA modes are needed to retain 99\% of the
cumulative variance for any of the four fields,
indicating that their dominant variations remain
concentrated in a small number of modes.

\begin{figure}[!htbp]
\centering
\includegraphics[width=0.92\linewidth]{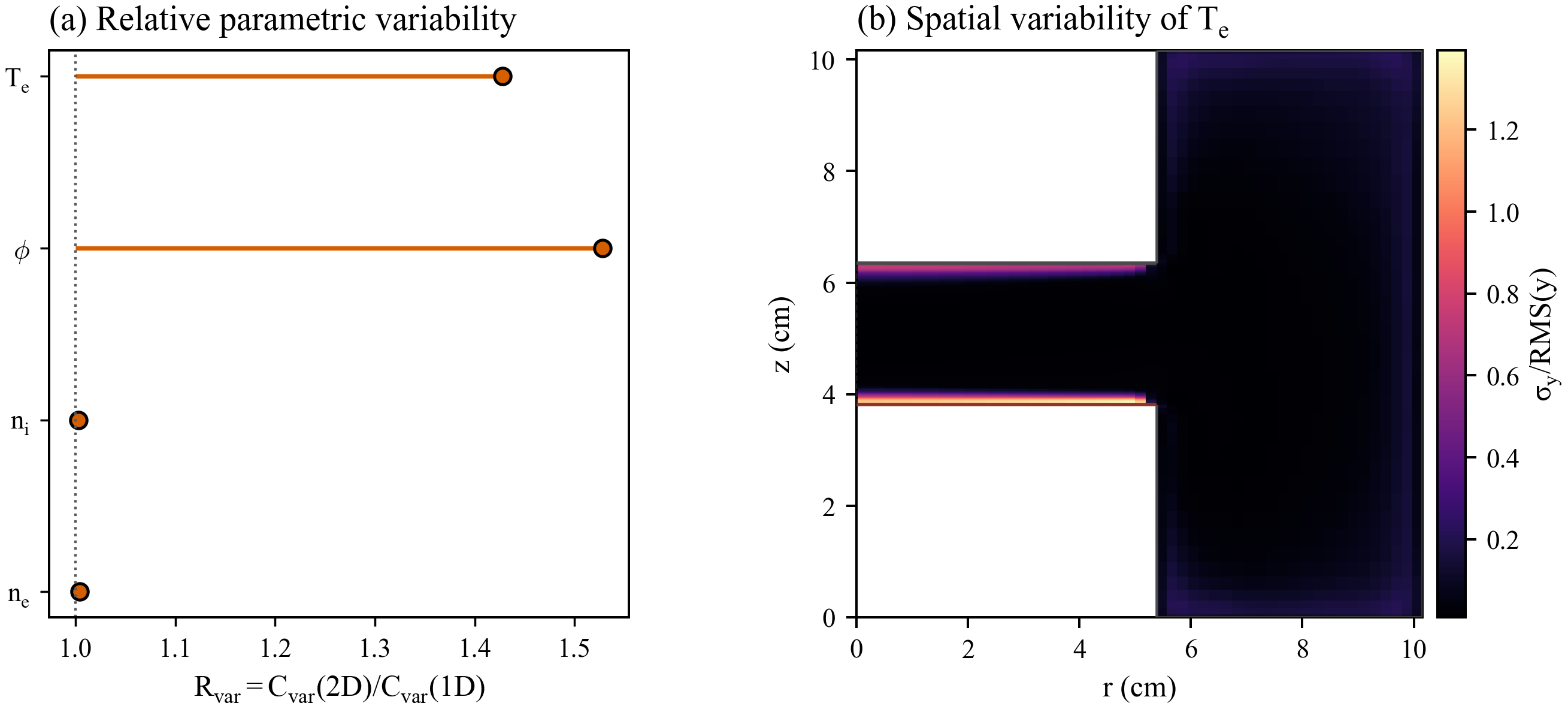}
\caption{Operating-condition-induced variability of the reference
CCP fields. (a) Ratios of normalized spatial variability,
$R_{\mathrm{var}}=C_{\mathrm{var}}^{2D}/C_{\mathrm{var}}^{1D}$,
for the four physical fields; stems begin at $R_{\mathrm{var}}=1$.
(b) Spatial distribution of normalized local $T_e$ variability
in the 2D GEC CCP. The density-field ratios are close to unity,
whereas those for electric potential and electron temperature
are 1.528 and 1.428, respectively.}
\label{fig:variability_final}
\end{figure}

The larger number of PCA modes required in 2D does not
imply a stronger response to operating conditions for
every field. Figure~\ref{fig:variability_final}(a)
compares the normalized variations across voltage,
pressure, and frequency in the two benchmarks.
The 2D-to-1D variability ratios are close to unity
for electron and ion densities, with values of
1.004 and 1.003, respectively, whereas the
corresponding ratios for $\phi$ and $T_e$ reach
1.528 and 1.428.

These differences may be associated with the electric-field
distribution and electron-energy transport in the 2D
chamber. The potential distribution is influenced by the
applied voltage and geometric boundaries, while the
central channel and radial expansion may lead to more
complex spatial redistribution of the electric field as
operating conditions vary. Electron temperature depends
on the balance between local electron heating, collisional
energy losses, and energy transport, and may therefore
respond more strongly to spatial variations in these
processes. The pronounced spatial nonuniformity of
the $T_e$ variability in
Fig.~\ref{fig:variability_final}(b) is consistent
with this interpretation. In contrast, electron and
ion densities are jointly influenced by ionization,
transport, and wall losses, and their normalized
variability does not exhibit a comparable increase.
These comparisons characterize the present two
benchmarks, which differ in their physical and
numerical settings as well as geometry.

\FloatBarrier

\subsection{Data-efficiency comparison}
\label{subsec:data_efficiency_v8}

\begin{figure}[!htbp]
\centering
\includegraphics[width=.92\linewidth]{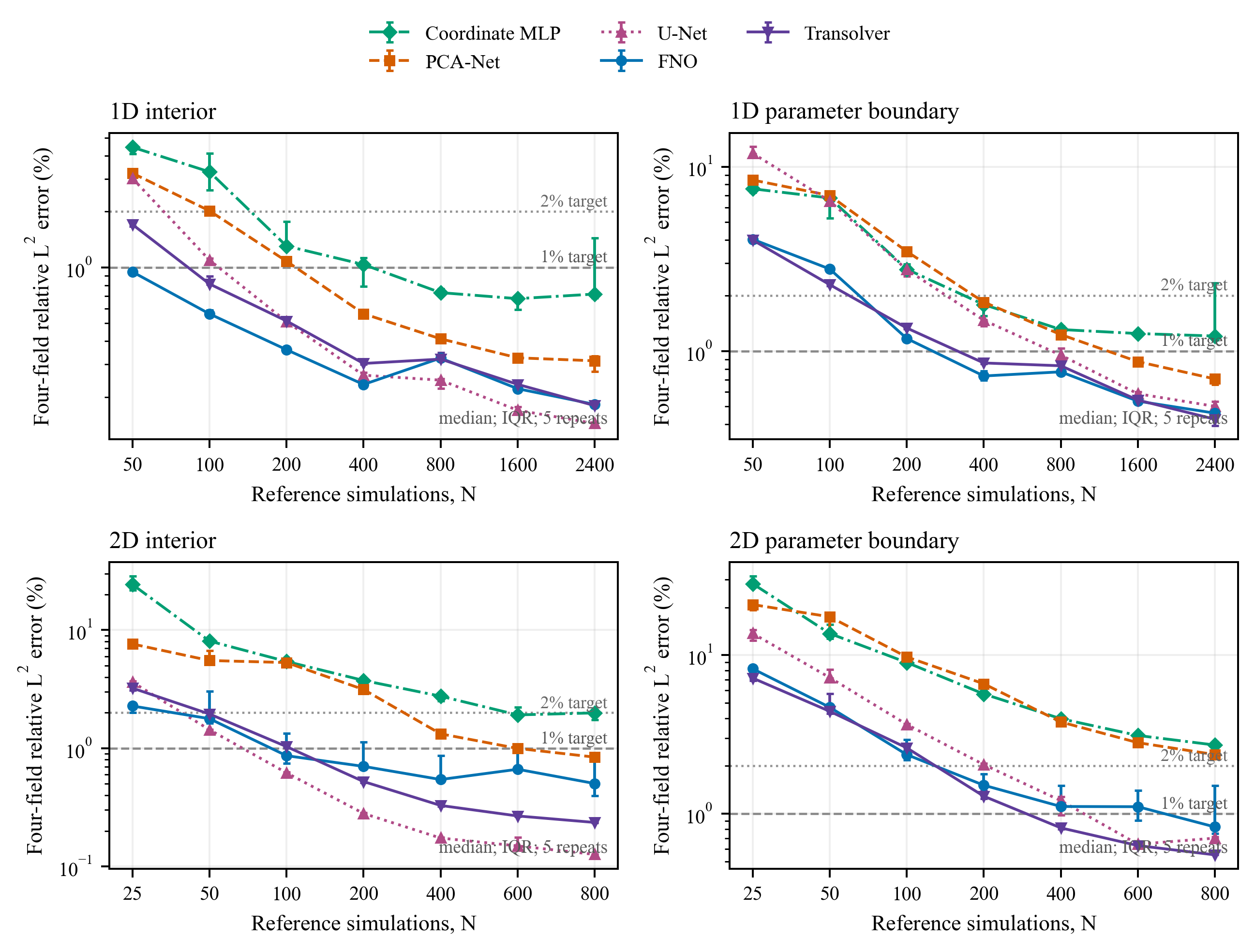}
\caption{Data-efficiency curves for the five surrogate models in the 1D
(top) and 2D (bottom) CCP benchmarks, evaluated on interior and
parameter-boundary test cases. Curves show five-repeat medians of
the four-field mean physical-space relative $L_2$ error, with Q1--Q3
error bars; horizontal lines indicate the 1\% and 2\% targets.
FNO requires the fewest or joint-fewest reference cases in seven
of the eight target comparisons (Table~\ref{tab:data_to_target_v8}).}
\label{fig:data_efficiency_v8}
\end{figure}

\paragraph{Reference-data requirements.}
Figure~\ref{fig:data_efficiency_v8} shows the five-model learning
curves, and Table~\ref{tab:data_to_target_v8} reports the minimum
tested training-set sizes required to meet the error targets.
FNO requires the lowest or joint-lowest number of reference cases
in seven of eight comparisons at the 1\% and 2\% targets.
Its advantage is more pronounced in the 1D benchmark: FNO reaches
1\% interior error with 50 cases, compared with 200 for U-Net.
In 2D, FNO gives the lowest median interior error at $N=25$ and
matches U-Net at the 1\% interior target (100 cases), while matching
Transolver at the 2\% parameter-boundary target (200 cases).
At the stricter 2D parameter-boundary 1\% target, however,
Transolver requires 400 cases compared with FNO's 800.
These results support FNO as a first-choice surrogate for the
CCP benchmarks studied here when reference simulations are limited,
while identifying higher-accuracy objectives for which U-Net and
Transolver offer stronger alternatives.

\begin{table}[!htbp]
\centering\small
\caption{Minimum tested reference cases needed to reach 2\%
and 1\% mean physical-space relative $L_2$ error. Each
threshold is determined from five-repeat medians; NR indicates
not reached. Smaller $N$ is better. The mean reference CPU
time per successful case is shown for context (single-core
runs; excluding unsuccessful attempts and retries).}
\label{tab:data_to_target_v8}
\setlength{\tabcolsep}{4pt}
\begin{tabular}{lrrrrrrrr}
\toprule
&\multicolumn{4}{c}{1D}&\multicolumn{4}{c}{2D}\\
\cmidrule(lr){2-5}\cmidrule(lr){6-9}
&\multicolumn{2}{c}{Interior}
&\multicolumn{2}{c}{Boundary}
&\multicolumn{2}{c}{Interior}
&\multicolumn{2}{c}{Boundary}\\
Model
&$N_{2\%}$&$N_{1\%}$
&$N_{2\%}$&$N_{1\%}$
&$N_{2\%}$&$N_{1\%}$
&$N_{2\%}$&$N_{1\%}$\\
\midrule
Coordinate MLP&200&800&400&NR&600&NR&NR&NR\\
PCA-Net&200&400&400&1600&400&600&NR&NR\\
U-Net&100&200&400&800&50&100&400&600\\
FNO&\textbf{50}&\textbf{50}&\textbf{200}&\textbf{400}
&\textbf{50}&\textbf{100}&\textbf{200}&800\\
Transolver&\textbf{50}&100&\textbf{200}&\textbf{400}
&\textbf{50}&200&\textbf{200}&\textbf{400}\\
\midrule
\shortstack[l]{Mean reference\\CPU time/case}
&\multicolumn{4}{c}{1.72 min}
&\multicolumn{4}{c}{3.80 h}\\
\bottomrule
\end{tabular}
\end{table}

\paragraph{Accuracy at the largest training-set sizes.}
Table~\ref{tab:main_performance_v8} reports the prediction errors
at the largest tested training-set sizes. U-Net achieves the
lowest interior errors in both 1D (0.144\%) and 2D
(0.126\%), while Transolver gives the lowest
parameter-boundary errors in 1D (0.427\%) and 2D
(0.550\%). These errors at the largest training-set sizes complement the
reference-case requirements reported in
Table~\ref{tab:data_to_target_v8}. Field-specific errors are
provided in Appendix
Figures~\ref{fig:s7_fieldwise_1d}
and~\ref{fig:s8_fieldwise_2d}.

\begin{table}[!htbp]
\centering\small
\caption{Errors at the maximum tested reference training-set
sizes ($N=2400$ in 1D; $N=800$ in 2D). Values are
percentages: median [Q1, Q3] over five repeats.}
\label{tab:main_performance_v8}
\setlength{\tabcolsep}{7pt}
\begin{tabular}{llrr}
\toprule
Benchmark&Model&Interior (\%)&Boundary (\%)\\
\midrule
\multirow{5}{*}{1D}
&Coordinate MLP&0.718 [0.685,1.437]&1.206 [1.149,2.339]\\
&PCA-Net&0.314 [0.274,0.334]&0.706 [0.658,0.746]\\
&U-Net&\textbf{0.144 [0.140,0.146]}&0.501 [0.490,0.533]\\
&FNO&0.183 [0.177,0.188]&0.461 [0.453,0.470]\\
&Transolver&0.180 [0.179,0.186]
&\textbf{0.427 [0.394,0.457]}\\
\midrule
\multirow{5}{*}{2D}
&Coordinate MLP&1.990 [1.748,2.124]&2.708 [2.620,2.847]\\
&PCA-Net&0.841 [0.827,0.851]&2.357 [2.216,2.364]\\
&U-Net&\textbf{0.126 [0.125,0.127]}&0.701 [0.694,0.713]\\
&FNO&0.503 [0.395,0.858]&0.824 [0.748,1.504]\\
&Transolver&0.235 [0.235,0.237]
&\textbf{0.550 [0.549,0.563]}\\
\bottomrule
\end{tabular}
\end{table}

\paragraph{Reference-simulation cost.}
The differences in reference-data requirements translate
directly into computational costs. Each accepted 2D CCP
reference simulation requires an average of 3.8024 CPU
core-hours (Table~\ref{tab:data_protocol}). At the 2D
parameter-boundary 2\% target, FNO requires 200 fewer
reference cases than U-Net, saving approximately 760 CPU
core-hours of reference generation. Tightening the FNO
target from 2\% to 1\%, however, increases the required
training set from 200 to 800 cases, a fourfold increase
corresponding to approximately 2281 additional CPU
core-hours. For 2D interior prediction, the same
improvement requires only 50 additional cases, or about
190 CPU core-hours. The additional reference-data cost
of stricter accuracy targets therefore depends strongly
on the prediction task, particularly near the boundary
of the operating-parameter range. Corresponding 1D
requirements for the 0.5\% target are provided in
Appendix Table~\ref{tab:appendix_halfpercent}.

At the 2D parameter-boundary 2\% target, FNO and
Transolver both require 200 reference cases. Their recorded
neural-network training times are listed in Appendix
Table~\ref{tab:training_costs}; however, the Transolver
cohort was timed under different execution conditions,
and the across-size medians do not permit a controlled
comparison of training speed. At the stricter 1\%
parameter-boundary target, Transolver requires 400 cases
rather than FNO's 800, corresponding to approximately
1521 fewer CPU core-hours of reference generation.
The principal computational comparison here therefore
concerns the reference-data requirements for the specified
error targets.

\begin{figure}[H]
\centering
\includegraphics[width=.75\linewidth]{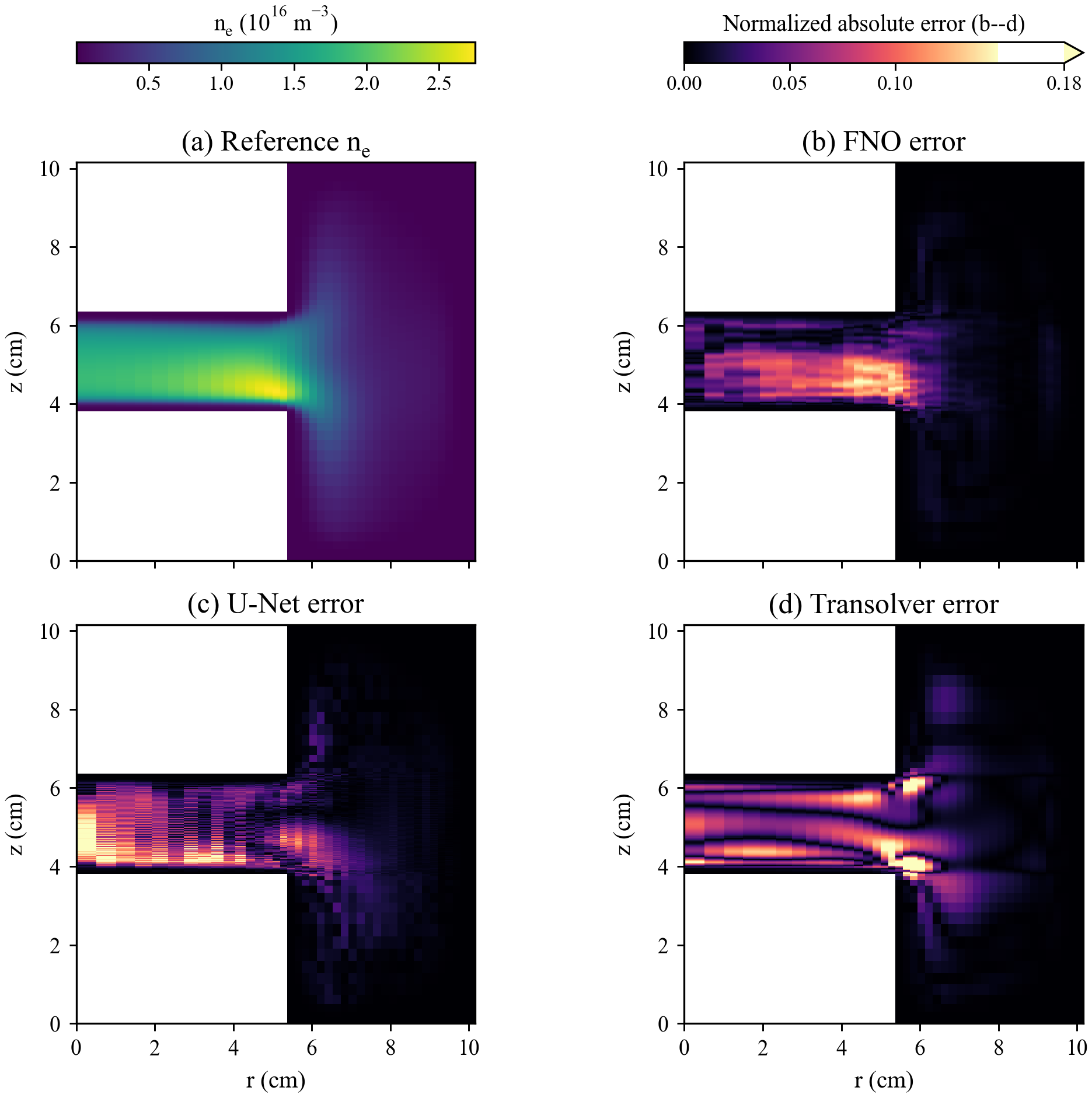}
\caption{Spatial electron-density prediction errors for a representative
2D interior case at $N=25$ (repeat 2): (a) reference $n_e$
(in $10^{16}\,\mathrm{m}^{-3}$) and (b)--(d) normalized absolute
errors for FNO, U-Net, and Transolver. The common error scale shows
$|\widehat n_e-n_e|/\mathrm{RMS}_{\mathrm{active}}(n_e)$,
with its upper limit set at the pooled active-cell 99th percentile.
The case and repeat were selected by proximity to the median paired
error across five repeats. The models exhibit distinct
channel and channel-exit error patterns.}
\label{fig:spatial_error_maps}
\end{figure}

\begin{figure}[H]
\centering
\includegraphics[width=.75\linewidth]{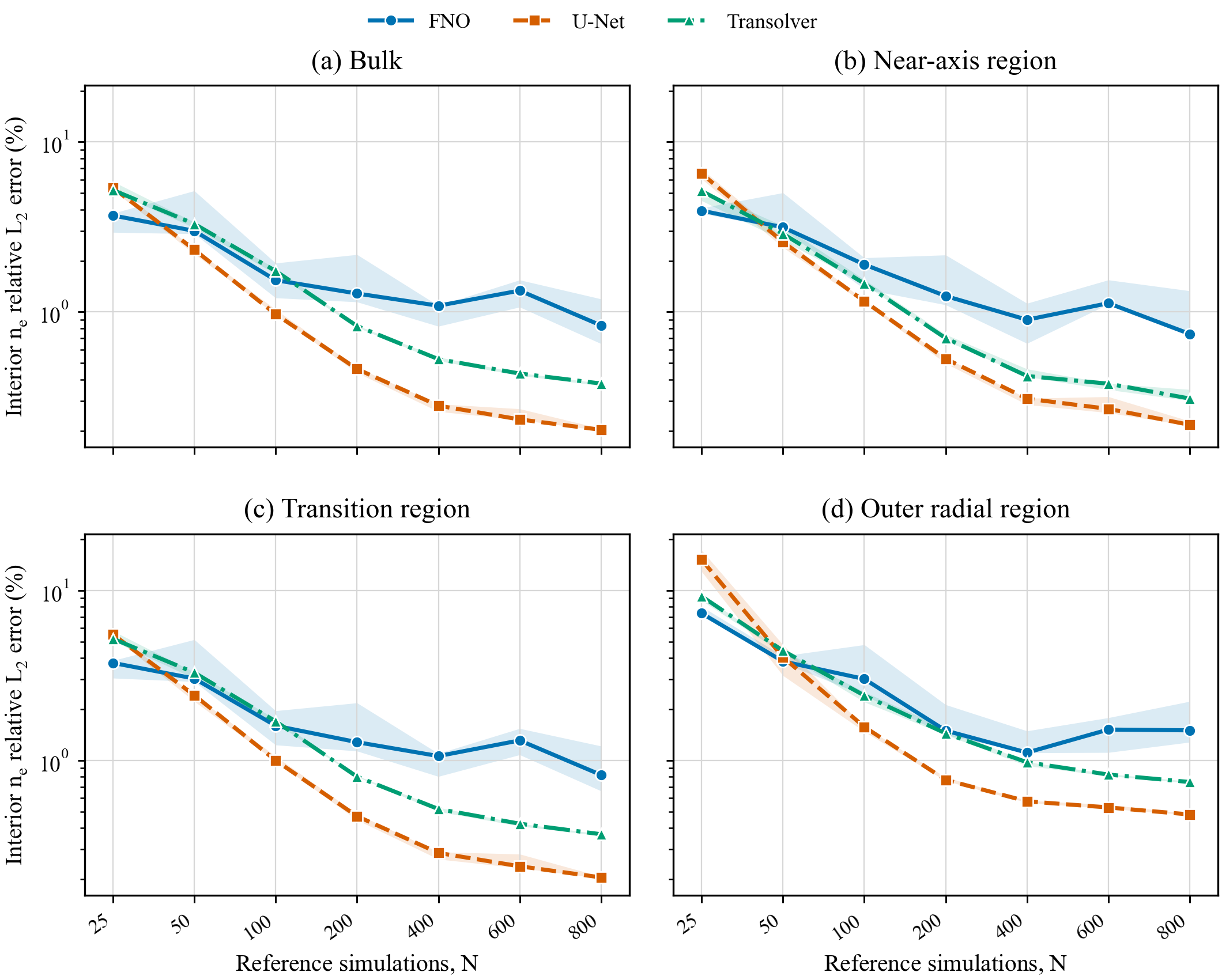}
\caption{Region-wise electron-density data efficiency over 100 held-out
2D interior test cases. Panels (a)--(d) show the bulk, near-axis,
transition, and outer-radial regions. Errors are regional
physical-space relative $L_2$ errors averaged over test cases
within each repeat; curves show five-repeat medians and Q1--Q3
intervals. FNO attains the lowest error in all four regions at
$N=25$, whereas U-Net does so at $N=200$.}
\label{fig:regional_data_efficiency}
\end{figure}

\paragraph{Spatial distribution of prediction errors.}
Figure~\ref{fig:spatial_error_maps} compares the electron-density
errors of FNO, U-Net, and Transolver for a representative
2D interior case at $N=25$, using a common normalized
absolute-error scale. FNO errors are concentrated mainly
within the high-density channel and near its exit, whereas
U-Net exhibits broader high-error regions within the narrow
channel. Transolver shows distinct band-like errors along
the channel and localized peaks near the geometric transition.
These spatial patterns identify the channel and its geometric
transition as informative regions for examining CCP field
reconstruction, complementing the region-wise learning curves
in Fig.~\ref{fig:regional_data_efficiency}.
\FloatBarrier

Figure~\ref{fig:regional_data_efficiency} further
quantifies electron-density data efficiency in four
geometric regions of the 2D GEC chamber. At the smallest tested
training-set size ($N=25$), FNO achieves the lowest median
regional error among the three models in all four regions.
For example, in the bulk region, the FNO error is
approximately 3.7\%, compared with 5.4\% for U-Net
and 5.2\% for Transolver. In the outer-radial region,
FNO gives approximately 7.3\% error, compared with
15.2\% and 9.2\%, respectively. These regional results agree with the four-field mean
comparison at $N=25$.
As the number of reference cases increases, U-Net
shows a stronger reduction in regional errors and
achieves the lowest median error in all four regions
at $N=200$. The shift from FNO at $N=25$ to U-Net at
$N=200$ shows that the most accurate regional predictor
changes with the reference-data budget, motivating the
representation-specific analyses in the following section.

\FloatBarrier
\subsection{Representation-specific analyses}
\label{sec:mechanisms_v9}

The axisymmetric GEC CCP combines bulk-scale field variations
with sharper gradients near sheaths and geometric transitions.
We therefore examine three forms of spatial representation:
FNO spectral modes and boundary treatment, U-Net multiscale
features and skip connections, and Transolver adaptive slicing
and token-level attention. The controls distinguish which
architectural choices affect reconstruction under the tested
reference-data budgets. We then consider PCA-Net from a complementary perspective,
distinguishing low-rank field representation from prediction
at unseen operating conditions.

\subsubsection*{FNO: Global spectral representation}

FNO represents spatial variations through globally supported
Fourier modes. The relatively smooth bulk-scale structures
of the CCP fields can be represented by lower spatial
frequencies, whereas sharper gradients near the sheaths and
geometric transitions generally require a broader range of
spectral components. The retained-mode configuration may
therefore affect the reconstruction of these different
spatial scales~\cite{ref21,ref22}.

\begin{figure}[!htbp]
\centering
\includegraphics[width=.94\linewidth]{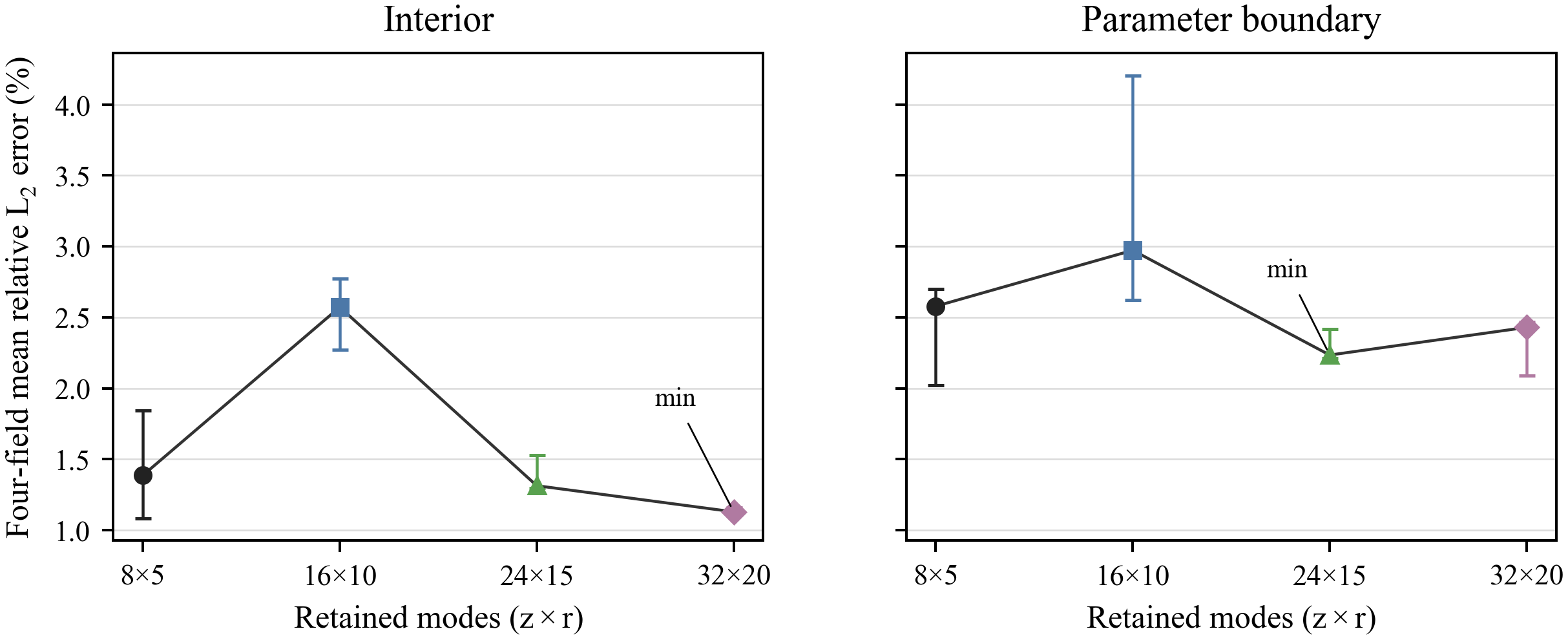}
\caption{Sensitivity of FNO prediction error to retained Fourier modes
at $N=800$, using near-parameter-matched configurations, 60 epochs,
and five repeats. The network width is adjusted across the four
mode settings; error bars show the median and Q1--Q3.
Increasing the number of retained modes does not monotonically
reduce error: the lowest interior error occurs at $32\times20$
modes, while the lowest parameter-boundary error occurs at
$24\times15$ modes.}
\label{fig:fno_retained_modes}
\end{figure}

Figure~\ref{fig:fno_retained_modes} compares four
near-parameter-matched configurations at $N=800$, retaining
$8\times5$, $16\times10$, $24\times15$, and $32\times20$
Fourier modes, with network widths adjusted to maintain
similar parameter counts. The prediction error does not
decrease monotonically with the number of retained modes.
For interior prediction, the $32\times20$ configuration
achieves the lowest median four-field mean relative
$L_2$ error of $1.128\%$, compared with $2.572\%$ for
$16\times10$. For parameter-boundary prediction, the
minimum error instead occurs at $24\times15$ ($2.234\%$),
slightly below that of $32\times20$ ($2.429\%$).
Thus, increasing spectral resolution does not necessarily
improve CCP field reconstruction; the combined choice
of retained modes and network width affects the
prediction accuracy. This 60-epoch mode-sensitivity study
is separate from the 1000-epoch formal FNO benchmark;
their absolute prediction errors are therefore not directly
comparable.

Spatial boundary treatment also matters: periodic RF excitation
in time does not make the CCP fields spatially periodic across
the chamber boundaries. Although FNO can learn non-periodic
solutions~\cite{ref22}, Fourier-based spatial extension can
introduce boundary mismatches. In the paired $N=200$ development
comparison, axial and outer-radial padding reduces median
regional errors for all four fields in each of the four
examined regions (Appendix Table~\ref{tab:padding_compact}).
The result supports treating spatial padding, alongside
spectral-mode selection, as an important configuration choice
for the fixed 2D GEC domain.

\FloatBarrier
\subsubsection*{U-Net: Local multiscale representation}

U-Net combines multiscale encoder--decoder features with
skip connections that transfer higher-resolution information
during reconstruction of CCP fields~\cite{ref23}.

\begin{figure}[!htbp]
\centering
\includegraphics[width=.45\linewidth]{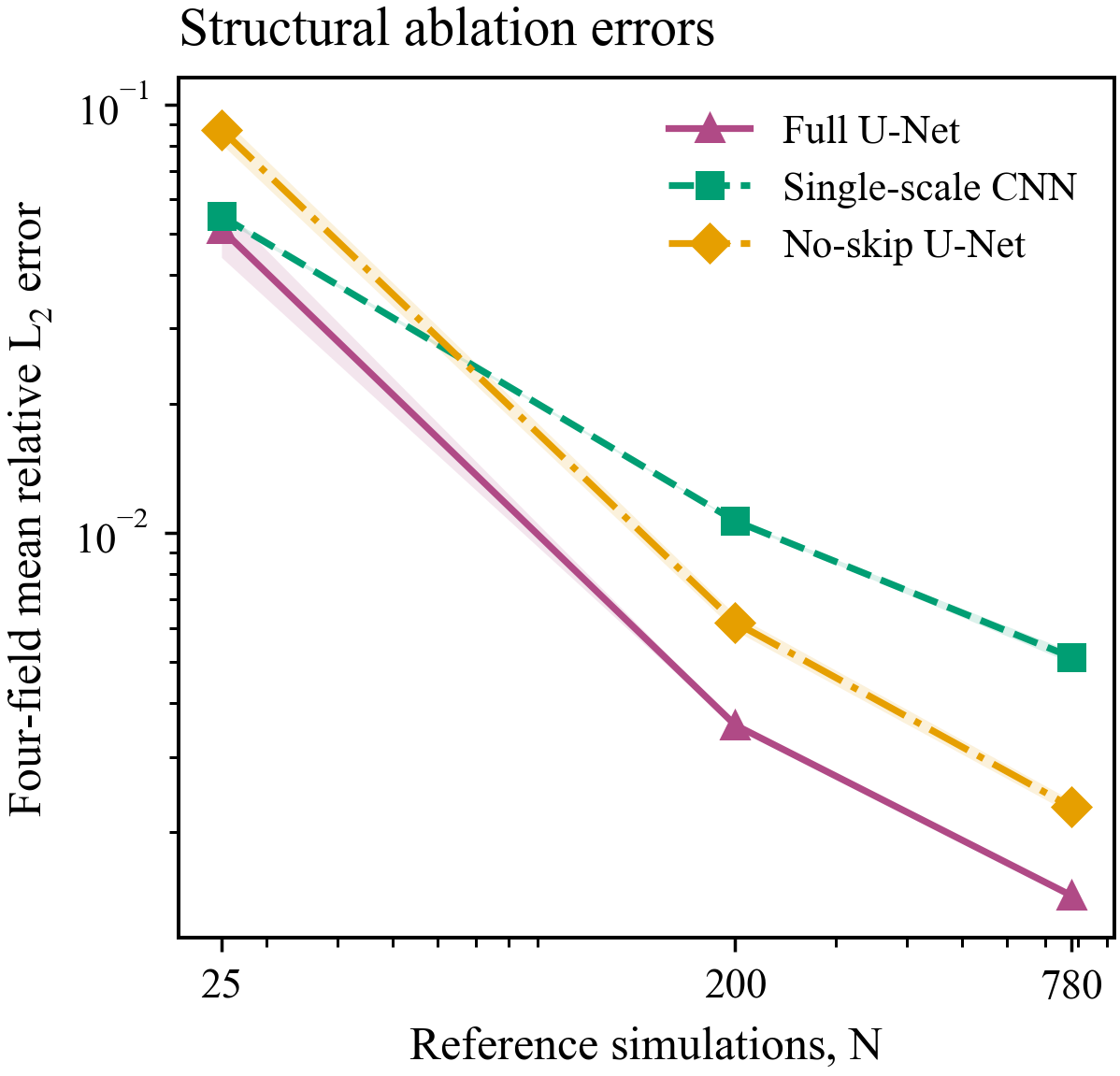}
\caption{U-Net structural controls comparing the full model with no-skip
and single-scale CNN variants at $N=25$, 200, and 780.
All configurations use a common 20-case development set,
two training repeats, and 1000 epochs; curves show the
mean four-field physical-space relative $L_2$ error and
shading spans the two repeat values. Full U-Net has the lowest
error at all three training-set sizes. The single-scale CNN
also differs in parameter count.}
\label{fig:unet_structural_final}
\end{figure}

Figure~\ref{fig:unet_structural_final} compares full U-Net,
a no-skip variant, and a single-scale CNN at $N=25$,
200, and 780. Full U-Net gives the lowest error at each
size, and removing skip connections consistently increases
error. The single-scale CNN performs relatively close to
full U-Net at $N=25$ but falls further behind at larger
budgets; because it also has fewer parameters, this
comparison does not isolate the multiscale hierarchy alone.

\begin{figure}[!htbp]
\centering
\includegraphics[width=.94\linewidth]{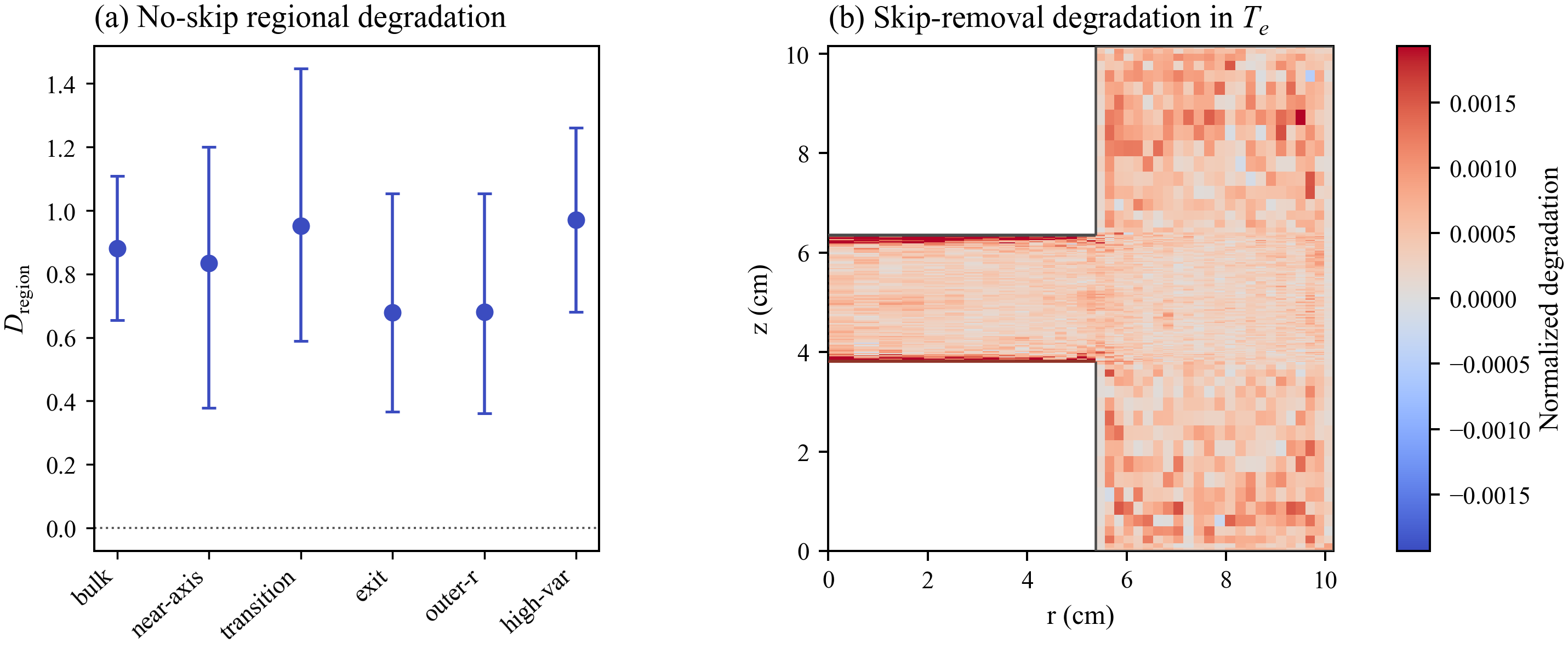}
\caption{Spatial effects of removing U-Net skip connections at $N=200$,
using 20 common development cases and two repeats.
(a) Regional $n_e$ error degradation (median and Q1--Q3);
(b) pointwise median normalized $T_e$ error degradation.
Positive values indicate higher errors without skip connections.
The degradation extends beyond localized gradients to multiple
regions of the plasma bulk and expanded chamber.}
\label{fig:unet_skip_spatial_final}
\end{figure}

The spatial comparison at $N=200$ in
Fig.~\ref{fig:unet_skip_spatial_final} shows that
removing skip connections increases $n_e$ errors in the
bulk, near-axis, transition, and expanded-chamber
regions. The corresponding $T_e$ error increase also
extends across multiple locations, rather than being
confined to narrow boundary-gradient zones.

To quantify the regional effect, Fig.~\ref{fig:unet_skip_spatial_final}(a) uses
\[
D_{\mathrm{region}}=
\frac{E_{\mathrm{no\mbox{-}skip}}-E_{\mathrm{full}}}
{\max(E_{\mathrm{full}},\epsilon)},
\qquad \epsilon=10^{-30},
\]
where $E_{\mathrm{no\mbox{-}skip}}$ and
$E_{\mathrm{full}}$ denote the regional RMS prediction
errors of the no-skip and full U-Net models, respectively,
after normalization by the corresponding case- and
field-specific reference RMS over active cells.
Positive $D_{\mathrm{region}}$ indicates increased error
after skip removal. Panel (a) reports the median and
Q1--Q3 for $n_e$ over 40 paired observations,
comprising 20 development cases and two repeats.

Figure~\ref{fig:unet_skip_spatial_final}(b) evaluates
the pointwise normalized absolute-error difference
for $T_e$,
\[
\frac{
|\widehat y_{\mathrm{no\mbox{-}skip}}-y|
-
|\widehat y_{\mathrm{full}}-y|
}{
\max[\RMS_{\Omega}(y),\epsilon]
},
\]
where $y$ is the reference electron-temperature field,
$\widehat y_{\mathrm{no\mbox{-}skip}}$ and
$\widehat y_{\mathrm{full}}$ are the corresponding
predictions, and $\RMS_{\Omega}(y)$ denotes the
reference-field RMS over the active domain $\Omega$.
The plotted value at each active cell is the median
over the same 40 paired observations; positive values
indicate larger local errors after skip removal.

Together, these controls associate U-Net skip connections
with improved field reconstruction across multiple plasma
regions, consistent with the regional trends in
Fig.~\ref{fig:regional_data_efficiency}.

\FloatBarrier
\subsubsection*{Transolver: Adaptive nonlocal representation}

In the 2D GEC CCP, locations with related field structures
need not lie in the same radial layer. Transolver addresses
such nonlocal interactions through input-dependent spatial
slicing rather than a prescribed radial partition.

\begin{figure}[!htbp]
\centering
\includegraphics[width=.94\linewidth]{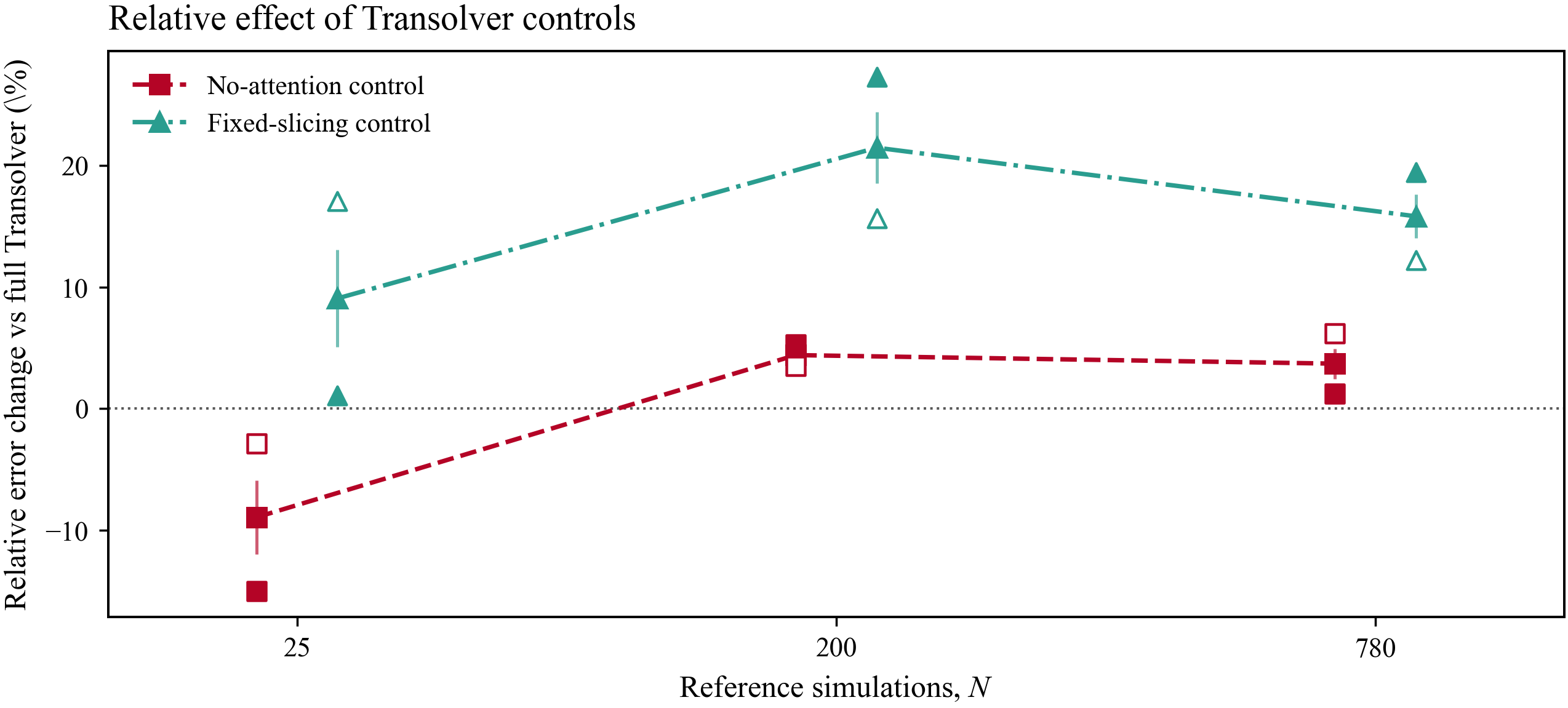}
\caption{Transolver structural controls comparing the full model,
no-attention, and fixed-radial-slicing variants at
$N=25$, 200, and 780, using 20 common development cases,
two repeats, and 1000 epochs. Plotted values are two-repeat
means of the relative error change $D_{\mathrm{control}}$;
positive values indicate higher error than the full model.
Fixed slicing increases error at all three sizes, whereas
removing QK attention has mixed effects. The no-attention
variant retains learned slicing and value/global-mean exchange.}
\label{fig:transolver_controls_final}
\end{figure}

Figure~\ref{fig:transolver_controls_final} compares full
Transolver with two controls: no-attention retains learned
slicing and token-level value exchange but removes
query--key (QK) attention, whereas fixed-radial slicing
retains attention but replaces the learned grouping.
At $N=25$, 200, and 780, removing QK attention changes
error by $-8.94\%$, $+4.41\%$, and $+3.70\%$,
respectively; its effect is therefore not uniformly
beneficial. In contrast, fixed-radial slicing increases
error by $9.07\%$, $21.47\%$, and $15.83\%$ at
the same budgets. Adaptive slicing has the more consistent
benefit in these controls, in line with analyses emphasizing
slicing and deslicing in Physics-Attention~\cite{hu2026transolverlinear}.

The relative error change for each control is defined as
\[
D_{\mathrm{control}}=
\frac{E_{\mathrm{control}}-E_{\mathrm{full}}}
{E_{\mathrm{full}}},
\]
where $E_{\mathrm{control}}$ and $E_{\mathrm{full}}$
are the errors of the control and full Transolver
models, respectively. Each error is averaged over
the 20 development cases and the four field-wise
physical-space relative $L_2$ errors. Positive
$D_{\mathrm{control}}$ indicates higher error for
the control model. The reported percentages are
the arithmetic means of the two paired repeat-level
relative changes at each training-set size.

\begin{figure}[!htbp]
\centering
\includegraphics[width=.94\linewidth]{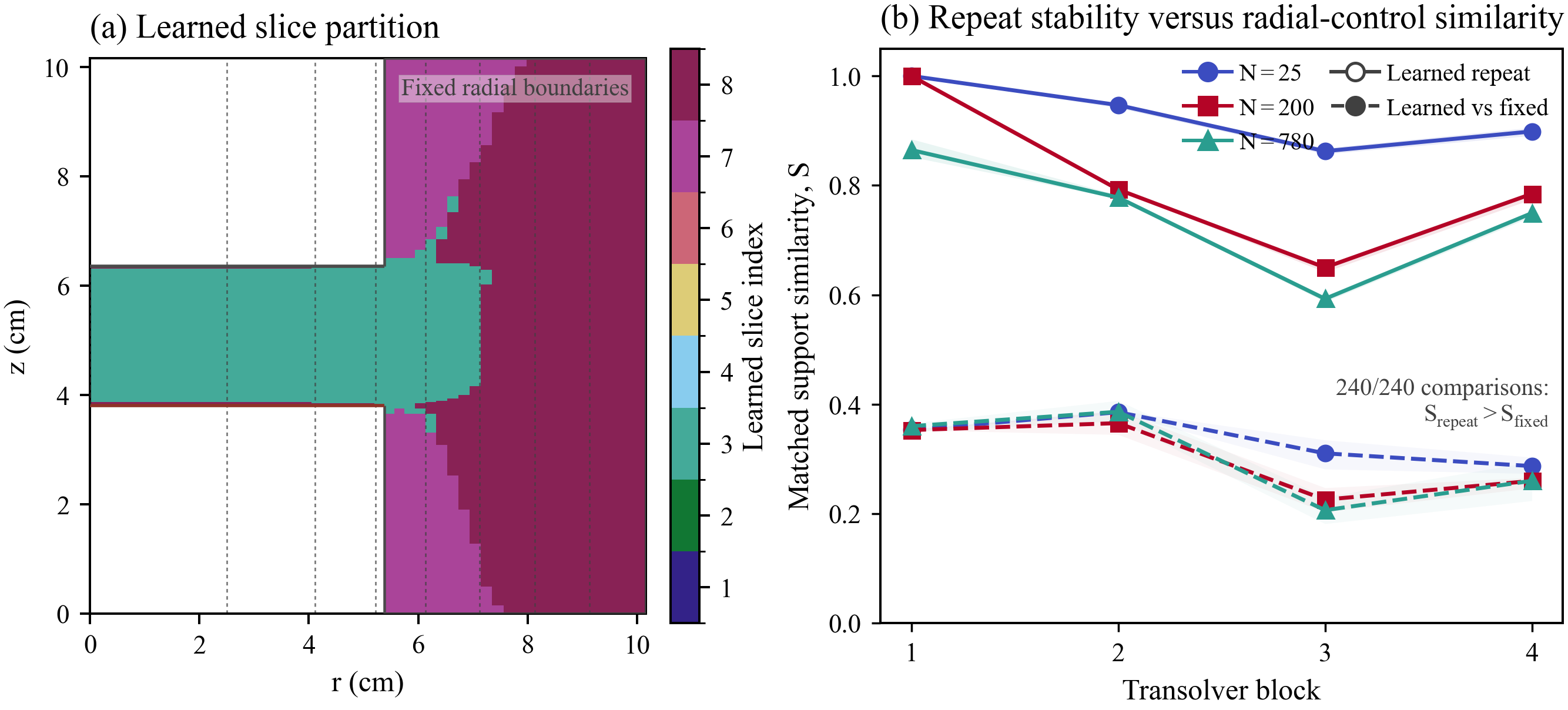}
\caption{Spatial organization and repeat stability of Transolver's
learned slicing. (a) Dominant slice assignments across the 2D
GEC CCP domain, with dotted lines indicating fixed radial
boundaries. (b) Matched support similarity between two training
repeats (solid lines) and between learned and fixed-radial slicing
(dashed lines), across four Transolver blocks at $N=25$, 200,
and 780. All 240 matched support comparisons (three training
sizes, 20 development cases, and four blocks) satisfy
$S_{\mathrm{repeat}}>S_{\mathrm{fixed}}$.}
\label{fig:transolver_slicing_final}
\end{figure}

Figure~\ref{fig:transolver_slicing_final}(a) shows that
learned slice groups extend across axial and radial
directions and are not restricted to fixed radial layers.
The groups span structures associated with the channel,
its exit, and the expanded chamber. Panel (b) compares
matched slice-support similarities: $S_{\mathrm{repeat}}$
between two training repeats and $S_{\mathrm{fixed}}$
between learned slicing and the prescribed radial partition
(see Appendix~\ref{app:mechanisms}).

All 240 training-size--case--block comparisons satisfy
\[
S_{\mathrm{repeat}}>S_{\mathrm{fixed}}.
\]
Across the two training repeats, the learned supports show
greater agreement with one another than with the fixed radial
partition. This repeat consistency, together with the reduced
prediction errors of adaptive slicing, supports its role in
representing nonlocal spatial organization across the CCP domain.

\FloatBarrier

\subsubsection*{PCA-Net: Representation versus prediction}
\label{sec:dimension_complexity_v9}

The preceding analyses examine spatially coupled architectures,
whereas PCA-Net provides a complementary perspective through
global low-rank reconstruction. As shown by the reference-field
PCA spectra in Sec.~\ref{subsec:reference_ccp_characteristics},
the dominant spatial variations can be represented by relatively
few modes. However, the ability to represent reference fields
with a low-dimensional basis does not directly determine
PCA-Net prediction accuracy at unseen operating conditions.
We therefore distinguish PCA reconstruction errors from
the errors arising in PCA-Net field prediction.

\begin{figure}[!htbp]
\centering
\includegraphics[width=0.94\linewidth]{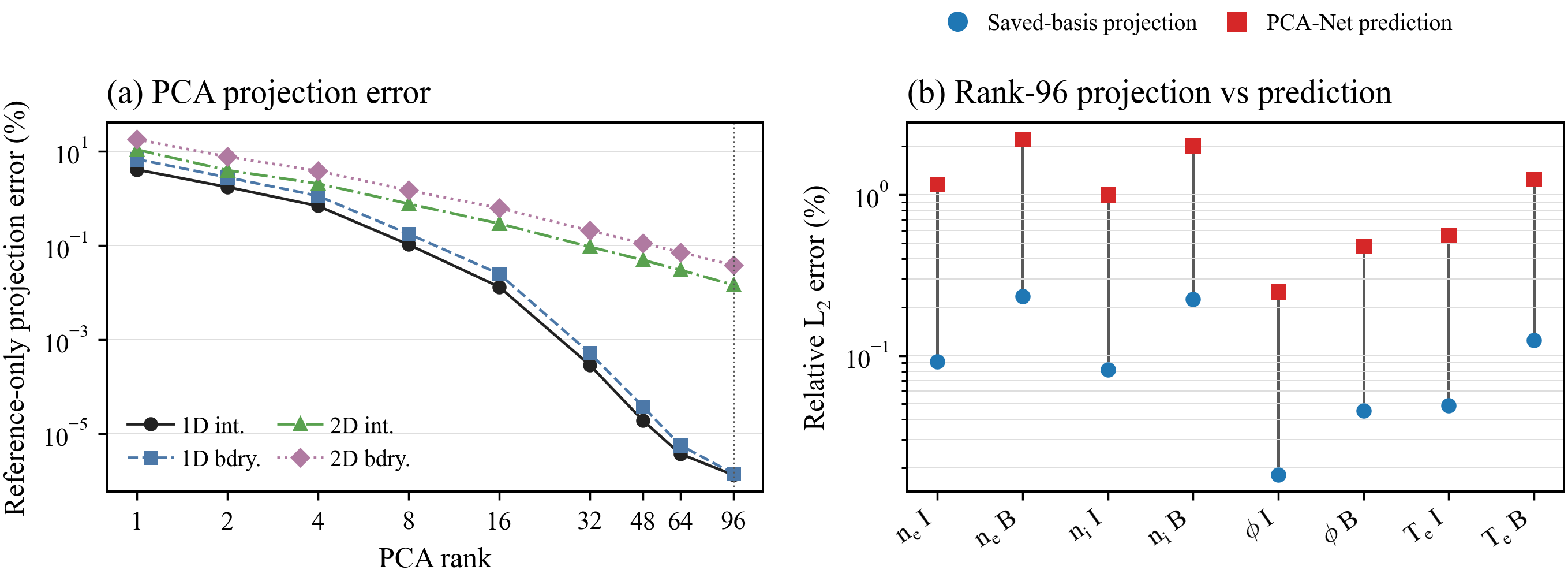}
\caption{PCA representation and prediction errors.
(a) Reference-only projection errors versus PCA rank on
held-out 1D and 2D interior and parameter-boundary test sets,
using PCA bases fitted to their respective training pools.
(b) Errors from projections onto each checkpoint's saved
rank-96 joint normalized basis, compared with PCA-Net
predictions in physical units. PCA-Net prediction errors
exceed the corresponding saved-basis projection errors by
factors of 9.78--13.53, indicating that basis representation
alone does not determine prediction accuracy.
The saved-basis projections are not independent per-field
physical-space optima; I and B denote interior and
parameter-boundary cases.}
\label{fig:pca_rank96_matched}
\end{figure}

To examine how accurately the reference fields can be
reconstructed from a limited number of PCA modes,
Fig.~\ref{fig:pca_rank96_matched}(a) compares the
projection errors of the 1D and 2D fields as the
PCA rank increases. Over the common rank range
$r_{\mathrm{PCA}}=1$--32, the 2D fields consistently
exhibit higher projection errors than their 1D
counterparts in both the interior and parameter-boundary
test sets. This further indicates that more PCA modes
are required to achieve comparable reconstruction
accuracy in the present 2D benchmark. The projection
sweep uses field-wise PCA bases fitted only on the
training pools and is distinct from the saved joint
normalized basis used by PCA-Net.

However, accurate PCA reconstruction of known fields
does not necessarily translate into accurate PCA-Net
predictions at unseen operating conditions.
Figure~\ref{fig:pca_rank96_matched}(b) compares the
projection errors of the saved rank-96 PCA basis
with PCA-Net prediction errors at $N=800$.
Across the eight field--test-set combinations,
the median matched prediction-to-projection error
ratios range from approximately 9.78 to 13.53.
This substantial gap shows that rank-96 truncation
alone cannot account for the final PCA-Net errors.
Although the PCA basis can reconstruct the reference
fields with relatively small errors, PCA-Net must
also predict the corresponding modal coefficients
from voltage, pressure, and frequency, as in non-intrusive
POD-based regression~\cite{ref19}. The accuracy
of this mapping therefore contributes to the
final field-prediction error. This distinction between PCA
representation and the complexity of learning the parameter-to-output
mapping is also reflected in theoretical analyses of
PCA-Net~\cite{lanthaler2023pcanet}.

\FloatBarrier
\subsection{Practical implications for surrogate selection}
\label{sec:selection_synthesis}

Together, the data-efficiency comparisons and
representation-specific analyses provide three practical
criteria for selecting CCP field surrogates. When reference CCP simulations are
limited, FNO is the preferred starting point, with its clearest
data-efficiency advantage in 1D and strong performance at the smallest
2D budgets. With larger training sets, U-Net provides the most accurate
interior-field predictions, while Transolver reaches the stringent 1\%
2D parameter-boundary target with 400 reference cases rather than
FNO's 800. Table~\ref{tab:surrogate_selection} summarizes these
choices in terms of reference-data budget, operating-parameter regime,
and accuracy target for the four cycle-averaged CCP fields.

\begin{table}[!htbp]
\centering
\small
\caption{Practical surrogate selection for the fixed-geometry CCP
benchmarks. Reference-data thresholds use the five-repeat median
four-field mean relative $L_2$ error; the errors at the largest tested training-set sizes are also
five-repeat medians.}
\label{tab:surrogate_selection}
\setlength{\tabcolsep}{3pt}
\renewcommand{\arraystretch}{1.15}
\begin{tabularx}{\linewidth}{@{}>{\raggedright\arraybackslash}p{0.30\linewidth}
>{\raggedright\arraybackslash}p{0.15\linewidth}
>{\raggedright\arraybackslash}X@{}}
\toprule
Application requirement & Preferred model & Supporting evidence \\
\midrule
Limited reference data &
\textbf{FNO} &
Lowest or joint-lowest reference-data requirement in seven of eight
1\%/2\% target comparisons. \\

High-accuracy interior-field prediction &
\textbf{U-Net} &
Lowest interior errors at the largest tested training-set sizes: 0.144\% (1D) and
0.126\% (2D); reaches the 2D interior 0.5\% target with 200 cases. \\

Stringent 2D parameter-boundary prediction &
\textbf{Transolver} &
Reaches the 1\% target with 400 rather than 800 cases for FNO,
saving approximately 1521 CPU core-hours of reference generation. \\
\bottomrule
\end{tabularx}
\end{table}

The appropriate accuracy measure can be refined further according
to the CCP physics of interest. For chamber-wide parameter scans,
the four-field domain-averaged error and the associated reference-data
requirement provide a useful initial comparison. For CCP studies
motivated by wafer etching, greater emphasis can be placed on the
spatial accuracy of ion density and electric potential near the
substrate-facing sheath, where charged-particle transport and
potential structure are particularly relevant to the wafer-facing
plasma environment~\cite{ref2,ref14}. Radially resolved errors
may also be important when discharge uniformity is the primary
objective. The channel and geometric-transition error patterns in
Fig.~\ref{fig:spatial_error_maps} illustrate how spatial diagnostics
can complement a global error measure. Table~\ref{tab:surrogate_selection}
therefore provides an initial model choice that can be refined using
field-specific and region-specific accuracy requirements.

\FloatBarrier
\section{Conclusion}

We compared five surrogate architectures for the prediction of
RF-cycle-averaged electron density, ion density, electric potential,
and electron temperature in 1D planar and 2D axisymmetric GEC CCP
benchmarks. By evaluating prescribed field-error targets over several
training-set sizes, we quantified reference-data requirements rather
than relying on accuracy at a single budget. FNO requires the fewest
or joint-fewest reference cases in seven of eight 1\% and 2\% target
comparisons, with its clearest advantage in 1D. In 2D, FNO and U-Net
both attain 1\% interior error with 100 cases. At the largest tested
training-set sizes, U-Net gives the lowest interior errors (0.144\%
in 1D and 0.126\% in 2D). For the stricter 2D parameter-boundary 1\%
target, Transolver requires 400 rather than FNO's 800 cases, equivalent
to approximately 1521 CPU core-hours of accepted-attempt reference
simulation time. These differences show why the preferred surrogate
depends on the prediction objective.

The spatial analyses reveal distinct error patterns within the
narrow discharge channel and near its geometric transition. FNO
accuracy depends on the retained spectral modes and spatial padding;
U-Net skip connections benefit reconstruction across several plasma
regions; and adaptive slicing provides a more consistent benefit in
the tested Transolver controls than query--key attention alone.
Reference-field PCA analysis further separates representational
capacity from predictive accuracy: in 2D, PCA-Net prediction errors
exceed the matched rank-96 basis projection errors by factors of
9.78--13.53. Thus, adequately representing the spatial fields does
not by itself ensure that their dependence on operating conditions
can be predicted accurately.

These findings provide a practical basis for matching surrogate
architecture to CCP modeling requirements. FNO is a preferred starting
point when the reference-simulation budget is limited; U-Net is favored
for high-accuracy interior fields with larger training sets, and
Transolver for stringent prediction near the 2D operating-parameter
boundary. In wafer-oriented applications, model assessment can place
additional weight on ion-density and potential reconstruction in
substrate-facing sheath regions, while studies of discharge uniformity
can emphasize radial and other spatially resolved errors. Extending
this reference-data-efficiency framework to further reactor geometries,
gas chemistries, and wafer-relevant plasma responses will help connect
surrogate selection more closely with the physical requirements of
plasma processing.

\section*{Acknowledgements}

This work was supported in part by Huawei Technologies Co., Ltd.
The authors acknowledge the computational resources and technical support
provided by the Theory Lab, Huawei Leibniz Research Center.

\section*{Competing interests}

Authors affiliated with Huawei Technologies Co., Ltd. include employees
and interns of the company. The authors declare no other competing
financial interests or personal relationships that could have influenced
the work reported in this paper.

\section*{Data availability}

The data that support the findings of this study are available from the
corresponding authors upon reasonable request.

\clearpage
\appendix
\setcounter{figure}{0}
\setcounter{table}{0}
\setcounter{equation}{0}
\renewcommand{\thefigure}{A\arabic{figure}}
\renewcommand{\thetable}{A\arabic{table}}
\renewcommand{\theequation}{A\arabic{equation}}

\section{Reference simulations and numerical checks}
\label{app:verification}

The reference CCP calculations solve electron and ion continuity, electron
energy, and Poisson equations as described in Sec.~2.1. Table~\ref{tab:solver_settings}
summarizes geometry-dependent modeling and numerical choices.
Scalar diagnostics and, in 2D, the cycle-averaged $T_e$ profiles
are sampled once per RF period and averaged over consecutive,
non-overlapping 20-cycle blocks. For a scalar diagnostic $D$,
the relative change between adjacent block averages is
\[
\eta_D=\frac{|D_b-D_{b-1}|}
{\max\left(|D_b|,|D_{b-1}|,10^{-30}\right)},
\]
where $b$ indexes the blocks. In 1D, spatial means are arithmetic averages,
and density maxima use instantaneous values at the end of each RF period.
In 2D, spatial means and the ion-density integral use axisymmetric cell-volume
weights over active cells, and density maxima use cycle-averaged fields.
The additional 2D electron-temperature profile metric is
\[
\eta_{T_e,\mathrm{prof}}=
\frac{\sqrt{\sum_{p\in\mathcal A}
[T_{e,b}(p)-T_{e,b-1}(p)]^2}}
{\max\left(\sqrt{\sum_{p\in\mathcal A}
\max\left(|T_{e,b}(p)|,|T_{e,b-1}(p)|\right)^2},10^{-30}\right)},
\]
where $\mathcal A$ is the active-cell set and $T_{e,b}(p)$ is the
block-averaged, cycle-averaged temperature. The profile metric does not
use cell-volume weighting.

\begin{table}[!htbp]
\centering
\footnotesize
\caption{Reference CCP model and numerical settings. Here $\gamma$ is the secondary-electron emission coefficient, and SparseLU denotes a sparse LU factorization.}
\label{tab:solver_settings}
\setlength{\tabcolsep}{4pt}
\begin{tabularx}{\linewidth}{@{}p{0.19\linewidth}YY@{}}
\toprule
Item & 1D planar CCP & 2D axisymmetric GEC CCP\\
\midrule
Geometry and grid & $25.4~\mathrm{mm}$ gap; 128 cells & $R=z=101.6~\mathrm{mm}$; powered radius $53.8~\mathrm{mm}$; $130\times40$ ($z\times r$) grid; 4624 active cells\\
Gas and species & Ar; $T_g=300~\mathrm{K}$; electrons and positive ions & Ar; $T_g=300~\mathrm{K}$; electrons and positive ions; no metastable branch\\
Reactions and energy & Tabulated direct ionization and excitation & Tabulated direct ionization and excitation; stepwise, superelastic, quenching and pooling channels disabled; ionization--excitation--elastic energy-loss model\\
Transport & Pressure-scaled constant mobility/diffusion & Pressure-scaled constant mobility/diffusion; electron-moment transport table\\
Wall treatment & Constant electron wall flux; $\gamma=0.01$ & 
Sakiyama-type electron/ion particle fluxes~\cite{sakiyama2006corona};
$\gamma=0.01$;
electron-energy wall flux uses adjacent-cell $T_e$;
fixed vessel-wall potential ($\phi_{\mathrm{wall}}=0$)
\\
Discretization and Poisson & Finite volume; upwind flux; semi-implicit update & Finite volume; upwind flux; semi-implicit Poisson update with electron-drift coupling; SparseLU, tolerance $10^{-10}$\\
Time stepping & Nominal 200 steps/RF; accepted retry at 400 steps/RF & 150 steps/RF\\
Convergence & Six scalar metrics; 20-cycle blocks; minimum 300 cycles; three checks below $5\times10^{-3}$ & Seven scalar metrics plus a $T_e$-profile metric; same stopping criterion\\
Supervision target & Average over final complete RF period & Average over final complete RF period\\
\bottomrule
\end{tabularx}
\end{table}

\begin{figure}[!htbp]
\centering
\includegraphics[width=.75\linewidth]{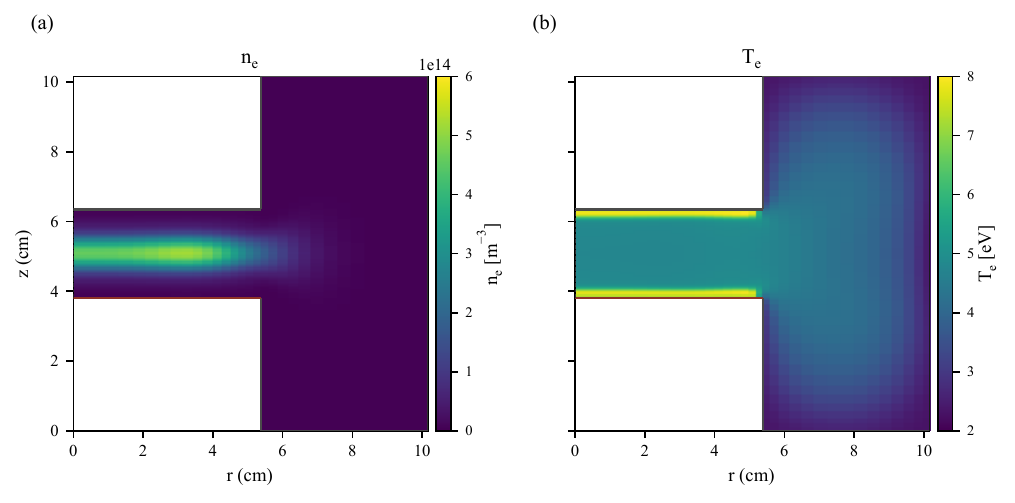}
\caption{Cycle-averaged electron-density $n_e$ and
electron-temperature $T_e$ distributions for the
2D axisymmetric GEC CCP solver check at
$P=100~\mathrm{mTorr}$ and $f=13.56~\mathrm{MHz}$.
The upper and lower electrodes are driven in
push--pull mode with RF amplitudes of $50~\mathrm{V}$
and a phase difference of $\pi$, with zero applied
DC offset.}
\label{fig:s2_solver_2d}
\end{figure}

Figure~\ref{fig:s2_solver_2d} presents the cycle-averaged
electron-density and electron-temperature distributions
obtained with the 2D axisymmetric GEC CCP solver for a
separate reference case at $P=100~\mathrm{mTorr}$ and
$f=13.56~\mathrm{MHz}$, outside the pressure range used
for surrogate training. The electrodes are driven in
push--pull mode with zero applied DC offset:
\[
V_{\mathrm{top}}(t)=50\sin(2\pi f t)~\mathrm{V},
\qquad
V_{\mathrm{bottom}}(t)=-50\sin(2\pi f t)~\mathrm{V}.
\]
The resulting spatial distributions are broadly consistent
with previously reported GEC CCP profiles~\cite{dechant2023zapdos,lymberopoulos1993_2d,lymberopoulos1994_modeling}.

\begin{figure}[!htbp]
\centering
\includegraphics[width=.75\linewidth]{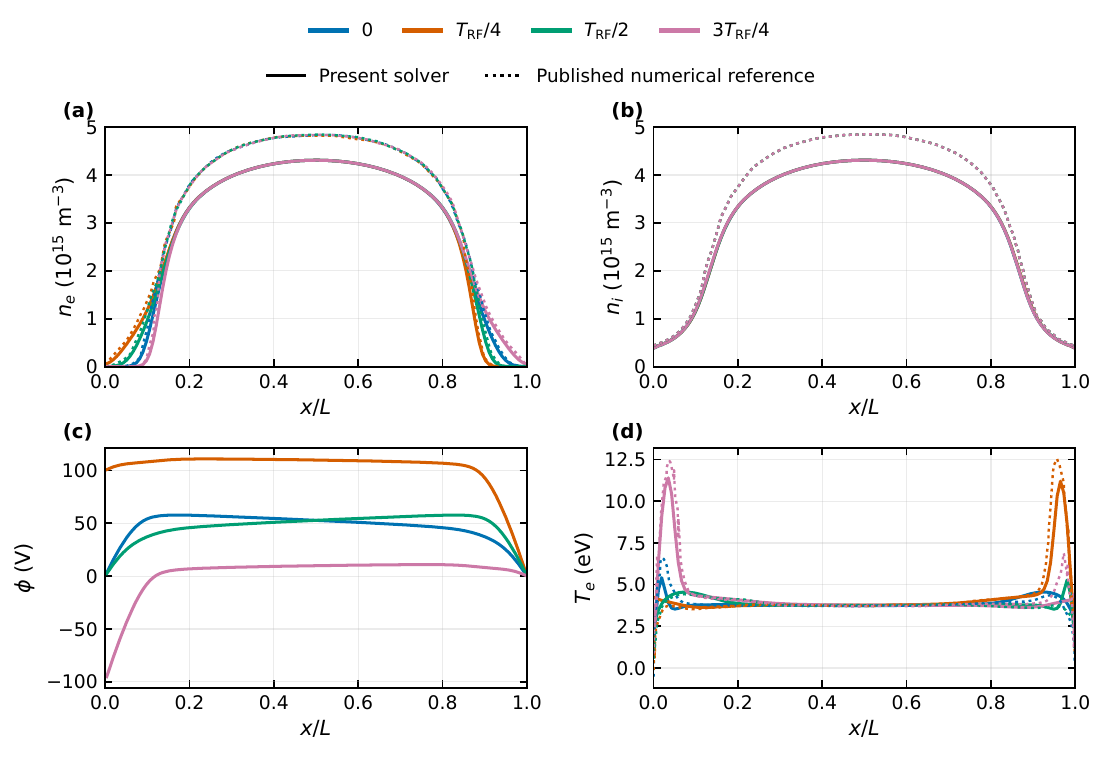}
\caption{Phase-resolved one-dimensional planar CCP fields at
$U_{\mathrm{RF}}=100~\mathrm{V}$, $P=1~\mathrm{Torr}$, and
$f=13.56~\mathrm{MHz}$. Solid curves are from the present
solver and dotted curves are the published reference profiles. The
published comparison contains $n_e$, $n_i$, and $T_e$; the $\phi$ panel shows
the present solver only.}
\label{fig:s1_solver_1d}
\end{figure}

Temporal and spatial sensitivities of each cycle-averaged reference field
$u$, distinct from surrogate errors, are evaluated at paired resolutions.
For spatial comparisons, the refined result is mapped onto the coarse grid:
\[
\delta_u=100\frac{\lVert u_{\mathrm{coarse}}-u_{\mathrm{refined}}\rVert_2}
{\lVert u_{\mathrm{refined}}\rVert_2}\%.
\]
Here $u_{\mathrm{coarse}}$ and $u_{\mathrm{refined}}$ are the corresponding
field vectors at the two compared resolutions.
\begin{table}[!htbp]
\centering
\small
\caption{Reference-solver temporal and spatial sensitivity (\%).
For both 1D and 2D, the low and center cases correspond to
$(U_{\mathrm{RF}},P,f)=(100~\mathrm{V},0.5~\mathrm{Torr},6.78~\mathrm{MHz})$
and $(150~\mathrm{V},1~\mathrm{Torr},27.12~\mathrm{MHz})$,
respectively. The high case corresponds to
$(200~\mathrm{V},2~\mathrm{Torr},54.24~\mathrm{MHz})$
in 1D and $(200~\mathrm{V},2~\mathrm{Torr},6.78~\mathrm{MHz})$
in 2D. Values compare the indicated numerical resolutions
and are not surrogate-prediction errors.}
\label{tab:numerical_sensitivity}
\setlength{\tabcolsep}{5pt}
\begin{tabular}{@{}llp{0.27\linewidth}rrrr@{}}
\toprule
System & Case & Comparison & $n_e$ & $n_i$ & $\phi$ & $T_e$\\
\midrule
1D & low & 200--400 steps/RF & 0.938 & 0.780 & 0.392 & 3.572\\
1D & center & 200--400 steps/RF & 0.237 & 0.230 & 1.069 & 5.721\\
1D & high & 200--400 steps/RF & 0.140 & 0.144 & 1.663 & 0.816\\
1D & center & 128--256 cells & 4.254 & 4.206 & 4.587 & 13.815\\
1D & center & 256--512 cells & 1.996 & 1.972 & 1.874 & 5.871\\
2D & low & 150--300 steps/RF & 0.348 & 0.391 & 1.193 & 0.919\\
2D & center & 150--300 steps/RF & 0.248 & 0.280 & 1.312 & 1.336\\
2D & high & 150--300 steps/RF & 0.197 & 0.267 & 1.398 & 7.127\\
2D & high & 300--600 steps/RF & 0.983 & 0.933 & 0.594 & 7.264\\
2D & center & $130\times40$ to $260\times80$ grid & 9.960 & 9.915 & 1.364 & 21.148\\
2D & center & $260\times80$ to $390\times120$ grid & 2.464 & 2.452 & 0.628 & 9.625\\
\bottomrule
\end{tabular}
\end{table}

\begin{figure}[!htbp]
\centering
\includegraphics[width=.96\linewidth]{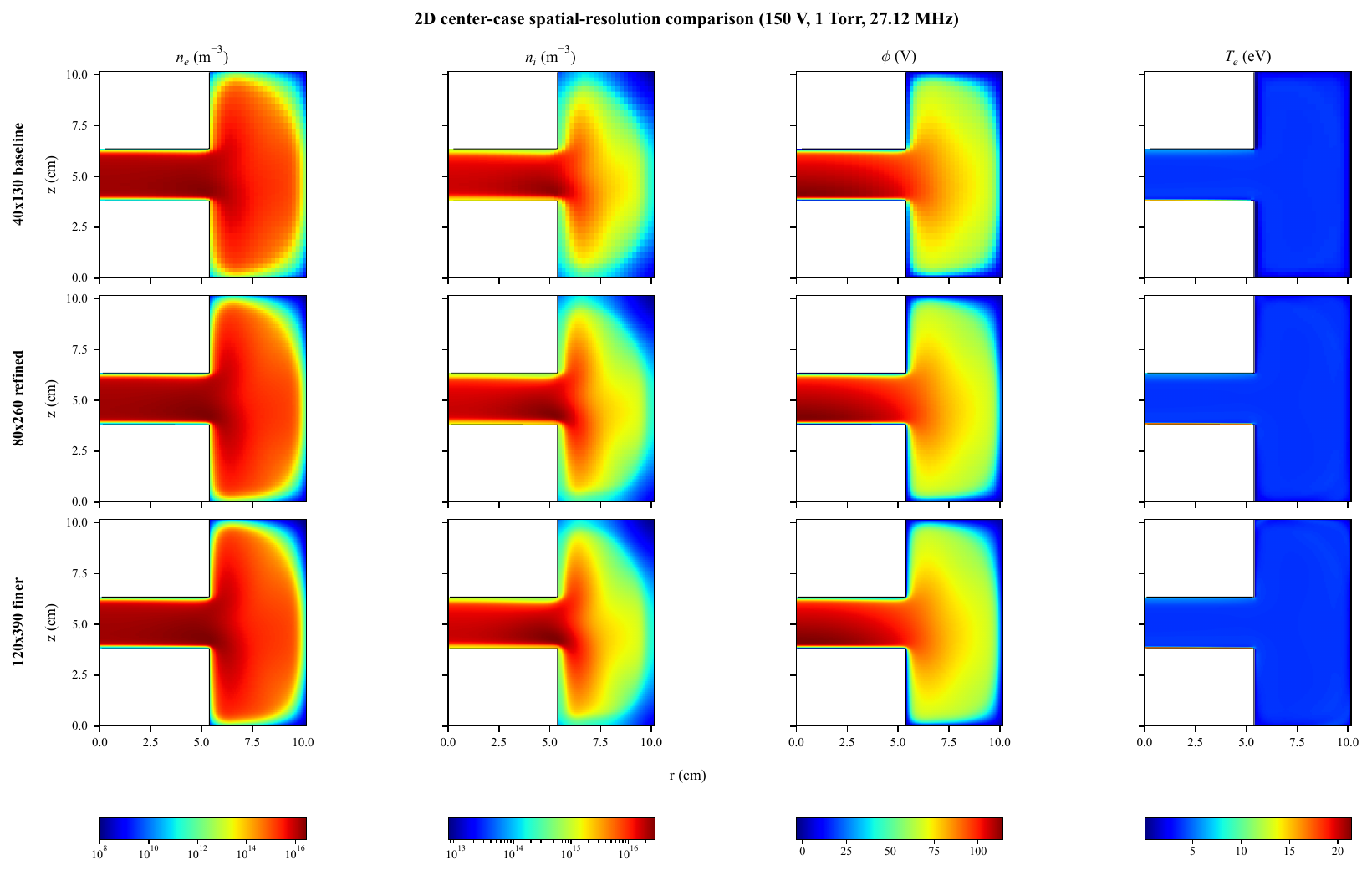}
\caption{Cycle-averaged fields for the 2D center case at the tested spatial
resolutions. Common color limits are used within each field; numerical
differences are listed in Table~\ref{tab:numerical_sensitivity}. Panel labels
show $r\times z$, while grid dimensions in the text and tables use $z\times r$.}
\label{fig:s3_resolution}
\end{figure}

Across the tested spatial resolutions, the principal
spatial structures of the 2D reference fields remain
qualitatively similar (Fig.~\ref{fig:s3_resolution}),
although quantitative differences persist,
particularly for $T_e$
(Table~\ref{tab:numerical_sensitivity}).
The surrogate errors in the main text are measured
relative to the specified reference grid, not to a
grid-independent plasma solution.

\FloatBarrier
\section{Dataset implementation and model development}
\label{app:data_methods}

The 1D reference set contains 2400 training-pool cases and two fixed test
sets of 300 cases each; the 2D set contains 800 training-pool cases and
100 interior plus 100 parameter-boundary test cases. The five architectures
share the training subset within each geometry, training-set size and repeat.
Operating parameters and output fields are normalized using training data
only. In 2D, Coordinate MLP samples 512 active spatial positions per selected
case for its training loss, while the grid-based models use the active grid.
Formal errors for all five models are evaluated on the same 4624 active cells;
outputs are returned to physical units before the relative $L_2$ evaluation.
For independent 2D structural studies, 20 cases from the training pool are
reserved for development, leaving up to $N=780$ fitting cases. This separate
development partition does not change the $N=800$ formal benchmark.

For the main benchmark, Coordinate MLP uses operating conditions and
spatial coordinates, PCA-Net maps operating conditions to modal coefficients,
and the remaining architectures take normalized operating parameters together
with their relevant grid, coordinate, or mask channels. In 2D, U-Net uses
coordinate channels and the active mask, FNO uses broadcast operating
parameters and the active mask, and Transolver operates on active spatial
points. All 2D reconstruction and evaluation exclude inactive cells.

The 1D axial extent is the coordinate containing 90\% of cumulative
positive electron-density content. In 2D the analogous radial extent
$r_{90}$ contains 90\% of the active-domain density integral
$\int_\Omega n_e\,\diff\mathcal V$, where $\diff\mathcal V=2\pi r\,\diff r\,\diff z$ is the cylindrical volume element.
The active-domain region masks used in FNO and U-Net tests are fixed before
examining model errors; they may overlap.

\subsection{Architecture-specific development settings}
Here LR denotes the learning rate, and OneCycleLR denotes a one-cycle
learning-rate schedule. Table~\ref{tab:tuning_candidates} summarizes
hyperparameter settings represented in the retained model-development
records; these settings were examined in selected combinations rather
than as an exhaustive search. 
Some settings were further refined through development-set
comparisons of training schedules and horizons before the
final configurations were adopted.
The adopted configurations for the main
comparisons are given in Table~\ref{tab:training_configs}.

\begin{table}[!htbp]
\centering
\caption{Selected architecture-specific hyperparameter settings
documented during model development. These settings do not
represent an exhaustive search or necessarily the final
benchmark configurations.}
\begingroup
\scriptsize
\setlength{\tabcolsep}{3.5pt}
\renewcommand{\arraystretch}{1.14}
\begin{tabularx}{\linewidth}{@{}p{0.13\linewidth}YY@{}}
\toprule
Architecture & Retained 1D candidates & Retained 2D candidates \\
\midrule
Coordinate MLP &
width $\{128,192,256\}$; depth $\{3,4\}$; epochs $\{120,240\}$;
LR $\{5\times10^{-4},10^{-3}\}$ &
width $\{192,256\}$; depth 4; epochs 240;
LR $5\times10^{-4}$; constant scheduler \\
PCA-Net &
rank $\{16,32,48,64\}$; coefficient width $\{96,128,192\}$;
epochs $\{120,240\}$; LR $\{5\times10^{-4},10^{-3}\}$ &
rank $\{96,128\}$; coefficient width 128; epochs 240;
LR $5\times10^{-4}$; constant scheduler \\
U-Net &
base width $\{16,32,64\}$; depth $\{3,4\}$; cap $\{128,256,512\}$;
epochs $\{120,240\}$; LR $\{5\times10^{-4},10^{-3}\}$;
constant/OneCycle pilots &
base width $\{48,64\}$; depth $\{3,4\}$ at width 64; cap 256;
epochs 1000; LR $5\times10^{-4}$; OneCycleLR \\
FNO &
corrected branch: width $\{64,128\}$; 3 layers; modes $\{32,64\}$;
padding $\{0,0.05,0.10,0.20\}$; constant-900/cosine-tail-200 &
width 64; 4 layers; modes $(32,20)$; padding variants F0--F3;
epochs 1000; LR $5\times10^{-4}$; constant/OneCycleLR \\
Transolver &
hidden width $\{128,192,256\}$; layers $\{4,5\}$; heads $\{8\}$;
slices $\{16,32\}$; epochs $\{60,120\}$; LR $5\times10^{-4}$;
constant/OneCycle pilots &
hidden width 128; 4 layers; 4 heads; slices $\{8,32\}$;
epochs 1000; LR $5\times10^{-4}$; OneCycleLR \\
\bottomrule
\end{tabularx}
\endgroup
\label{tab:tuning_candidates}
\end{table}

\begin{table}[!htbp]
\centering
\footnotesize
\caption{Surrogate architectures and adopted optimization settings. AdamW is used throughout. The 2D Coordinate MLP samples 512 active positions per selected case during training; formal evaluation for every model uses all 4624 active cells.}
\label{tab:training_configs}
\setlength{\tabcolsep}{3.8pt}
\begin{tabularx}{\linewidth}{@{}llYrrrl@{}}
\toprule
System & Model & Architecture & Epochs & LR & Batch & Scheduler\\
\midrule
1D & Coordinate MLP & Width 128; 3 hidden layers & 240 & 0.001 & 32 & constant\\
1D & PCA-Net & Rank 32; coefficient width 192 & 240 & 0.0005 & 32 & constant\\
1D & U-Net & Base 32; depth 3; cap 256 & 600 & 0.001 & 32 & OneCycleLR\\
1D & FNO & Width 128; 3 layers; 64 modes; padding 0.10 & 200 & 0.001 & 32 & cosine-tail\\
1D & Transolver & Width 192; 4 layers; 8 heads; 16 slices & 600 & 0.0005 & 16 & OneCycleLR\\
2D & Coordinate MLP & Width 256; 4 hidden layers & 240 & 0.0005 & 16 & constant\\
2D & PCA-Net & Target rank 96; width 128; effective rank $\min(96,N-1)$ & 240 & 0.0005 & 16 & constant\\
2D & U-Net & Base 64; depth 3; cap 256 & 1000 & 0.0005 & 16 & OneCycleLR\\
2D & FNO & Width 64; 4 layers; modes $(32,20)$; axial 0.10, outer-radial 0.05 padding & 1000 & 0.0005 & 16 & constant\\
2D & Transolver & Width 128; 4 layers; 4 heads; 8 slices & 1000 & 0.0005 & 2 & OneCycleLR\\
\bottomrule
\end{tabularx}
\end{table}

For the 2D networks, recorded parameter counts include approximately
$5.82\times 10^6$ for FNO and $0.513\times 10^6$ for Transolver; parameter count
alone therefore does not describe the observed elapsed training time.
The 2D PCA-Net effective rank is $\min(96,N-1,r_{\mathrm{available}})$,
where $r_{\mathrm{available}}$ is the number of fitted PCA components.
It equals 24, 49, and 96 at $N=25$, 50, and $N\ge100$, respectively.

\subsection{Training environment and recorded times}
The surrogate-training software environment comprised Python 3.10.14,
PyTorch 2.7.0, and NeuralOperator 2.0. Neural-network training used GPU
acceleration. The recorded training times characterize resource use
in the present experiments and should not be interpreted as
hardware-independent performance benchmarks.
Table~\ref{tab:training_costs} records neural-network training time
separately from CPU reference-simulation cost. Each row summarizes seven
training-set sizes and five repeats. The median is the median time per
completed training run across these sizes, whereas the sum is the
cumulative time for all 35 runs. The 2D Transolver cohort was recorded separately from the other models;
its different execution and timing conditions mean that the observations
should not be interpreted as a controlled same-$N$ training-speed benchmark.

\begin{table}[!htbp]
\centering
\small
\caption{Observed neural-network training times for 35 runs per model. Median h/run is taken across training-set sizes and repeats; summed GPU-h is the recorded total over 35 runs, not the cost of training one final surrogate. The 2D Transolver runs were timed separately.}
\label{tab:training_costs}
\setlength{\tabcolsep}{5pt}
\begin{tabular}{@{}llrrr@{}}
\toprule
System & Model & Runs & Median h/run & Summed GPU-h\\
\midrule
1D & Coordinate MLP & 35 & 0.0024 & 0.1502\\
1D & FNO & 35 & 0.0083 & 0.5564\\
1D & PCA-Net & 35 & 0.0019 & 0.1236\\
1D & Transolver & 35 & 0.1157 & 7.3290\\
1D & U-Net & 35 & 0.0335 & 2.1635\\
2D & Coordinate MLP & 35 & 0.1606 & 7.9337\\
2D & FNO & 35 & 0.5320 & 27.9949\\
2D & PCA-Net & 35 & 0.0240 & 1.1546\\
2D & Transolver & 35 & 1.6974 & 93.1407\\
2D & U-Net & 35 & 0.1645 & 8.9127\\
\bottomrule
\end{tabular}
\end{table}

\section{Additional data-efficiency results}

\begin{figure}[!htbp]
\centering
\includegraphics[width=.75\linewidth]{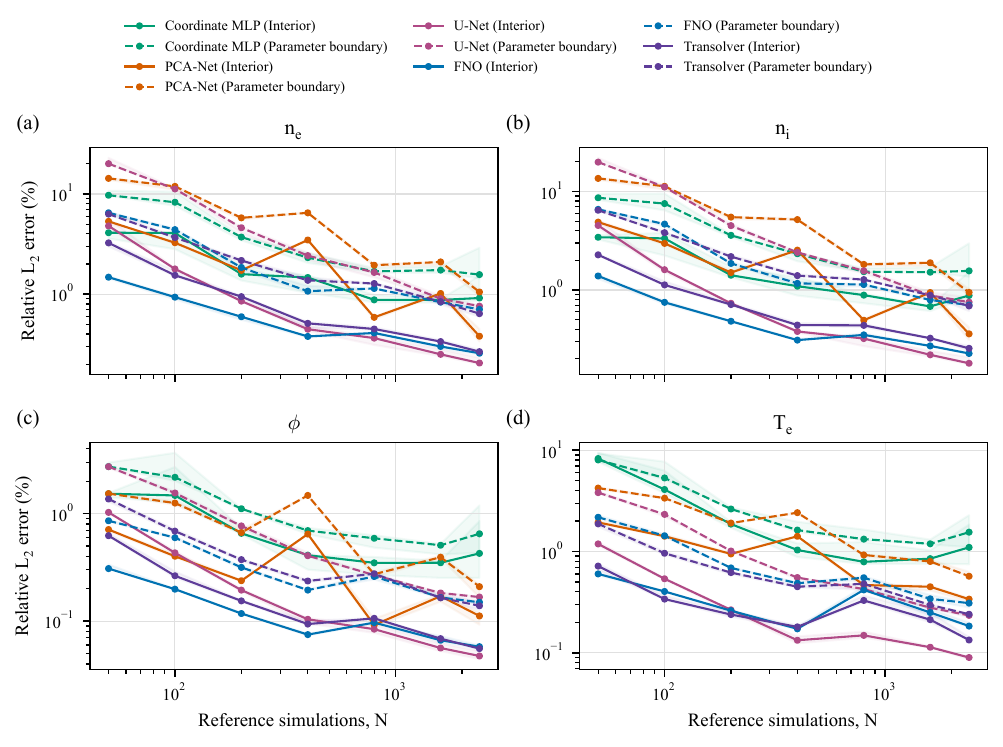}
\caption{Field-wise one-dimensional learning curves. Curves show medians
and shaded Q1--Q3 intervals over five repeats; solid and dashed lines denote
interior and parameter-boundary cases, respectively.}
\label{fig:s7_fieldwise_1d}
\end{figure}

\begin{figure}[!htbp]
\centering
\includegraphics[width=.75\linewidth]{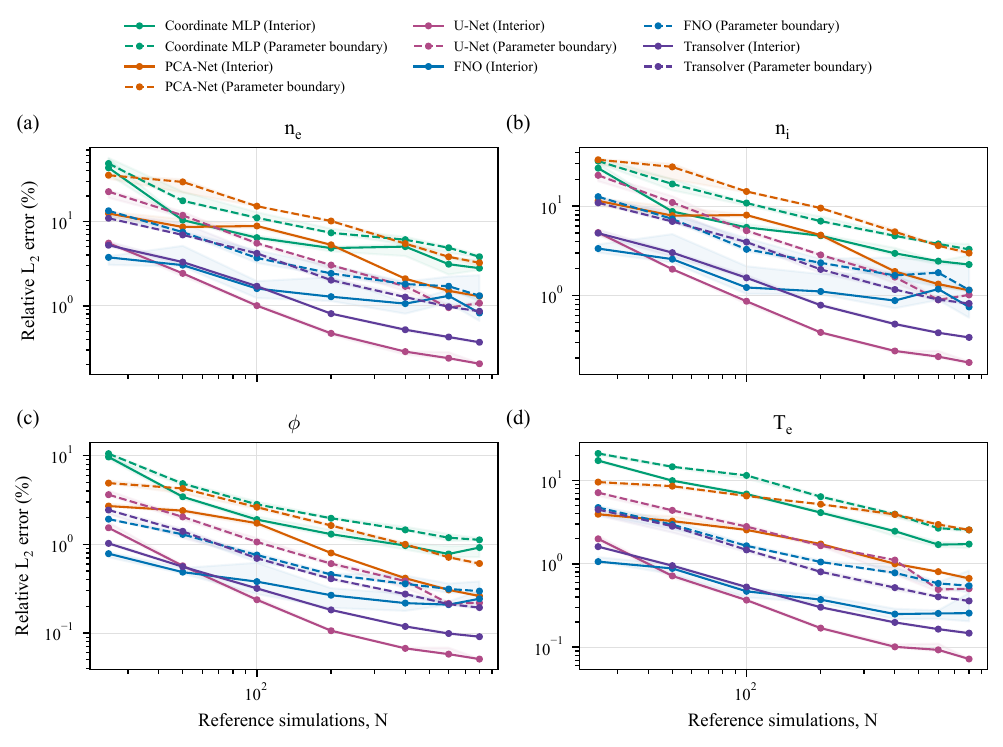}
\caption{Field-wise two-dimensional learning curves. Curves show medians
and shaded Q1--Q3 intervals over five repeats; solid and dashed lines denote
interior and parameter-boundary cases, respectively.}
\label{fig:s8_fieldwise_2d}
\end{figure}

The full-training-set five-repeat four-field mean threshold requirements at
2\% and 1\% are given in Table~\ref{tab:data_to_target_v8}.
Figures~\ref{fig:s7_fieldwise_1d} and~\ref{fig:s8_fieldwise_2d}
resolve the four predicted fields individually.

The more stringent 0.5\% threshold is reported separately to
show the added reference-data demand without repeating the 1\% and 2\%
values from the main text. NR denotes not reached at any tested training-set size.
\begin{table}[!htbp]
\centering\small
\caption{Minimum tested reference cases to attain a 0.5\% four-field mean
physical-space relative $L_2$ error. NR: not reached in the tested training-set sizes.}
\label{tab:appendix_halfpercent}
\begin{tabular}{@{}lrrrr@{}}\toprule
&\multicolumn{2}{c}{1D}&\multicolumn{2}{c}{2D}\\
\cmidrule(lr){2-3}\cmidrule(lr){4-5}
Model&Interior&Boundary&Interior&Boundary\\\midrule
Coordinate MLP&NR&NR&NR&NR\\
PCA-Net&800&NR&NR&NR\\
U-Net&400&NR&200&NR\\
FNO&200&2400&NR&NR\\
Transolver&400&2400&400&NR\\
\bottomrule\end{tabular}
\end{table}

\FloatBarrier
\section{Additional representation analyses}
\label{app:mechanisms}

\subsection{FNO spectral modes and spatial padding}
The independent 2D FNO mode study compares $(m_z,m_r)=(8,5),\,(16,10),\,
(24,15),\,(32,20)$ with approximately matched parameter counts at
$N=800$, using 60 training epochs and five repeats. Figure~\ref{fig:fno_retained_modes}
shows the corresponding four-field mean errors. The final formal benchmark
uses the separately selected 1000-epoch FNO configuration.

To examine the non-periodic GEC boundaries, the padding control compares
an unpadded configuration against axial padding of 10\% at both ends and outer-radial
padding of 5\% (no inner-radial padding) at $N=200$ on the common
20-case development set. Both use 1000 epochs and two repeats.
For each paired case, the normalized gain is
\[
G=\frac{E_{\mathrm{unpad}}-E_{\mathrm{pad}}}
{\max(E_{\mathrm{unpad}},\varepsilon_E)},
\]
where $E$ is the corresponding dimensionless regional relative error and
$\varepsilon_E=10^{-30}$ is a dimensionless numerical safeguard. Positive values
indicate improvement with padding. Table~\ref{tab:padding_compact}
shows the median regional gains in the four output fields.
\begin{table}[!htbp]
\centering\small
\caption{Median normalized regional error reduction with FNO boundary
padding, from the paired $N=200$ development comparison (two repeats).}
\label{tab:padding_compact}
\begin{tabular}{@{}lrrrr@{}}\toprule
Field&Bulk&Near-axis&Transition&Outer radial\\\midrule
$n_e$&0.393&0.391&0.478&0.358\\
$n_i$&0.396&0.477&0.503&0.313\\
$\phi$&0.293&0.353&0.332&0.187\\
$T_e$&0.389&0.498&0.284&0.193\\
\bottomrule\end{tabular}
\end{table}
The corresponding global potential-range descriptor exhibits a median
normalized gain of 0.678; the radial extent $r_{90}$ shows negligible median
change. Here near-axis denotes the innermost 12\% of the radial cell-center
extent. The regional values are based on predeclared masks, which may overlap.

\subsection{U-Net structural controls}
Full, no-skip and single-scale convolutional controls are compared at
$N=25,200,780$ on common development cases with two repeats. The
single-scale control has a different parameter count; these experiments
measure architecture sensitivity rather than a parameter-matched causal
isolation of multiscale depth. The main text reports the corresponding
performance and spatial-error results (Figs.~\ref{fig:unet_structural_final}
and~\ref{fig:unet_skip_spatial_final}).

For Fig.~\ref{fig:unet_skip_spatial_final}(a), the fixed masks are bulk
($0.12<\widetilde r<0.82$, $0.12<\widetilde z<0.88$), near-axis
($\widetilde r\leq0.12$), transition
($0.35\leq\widetilde z\leq0.70$), channel-exit expansion
($\widetilde z\geq0.70$), outer radial ($\widetilde r\geq0.82$), and
high-variability (the transition band intersected with
$\widetilde r\geq0.45$), where $\widetilde r=(r-r_{\min})/(r_{\max}-r_{\min})$
and $\widetilde z=(z-z_{\min})/(z_{\max}-z_{\min})$ use the active-grid cell-center
extents. Thus, the near-axis mask touches the symmetry axis rather than a
material wall. The masks may overlap.

\subsection{Transolver slicing controls}
The full, no-attention and fixed-radial-slicing variants use the same three
development training-set sizes and two repeats. Each of the four blocks contains
four heads and eight slices per head. Comparing learned slicing supports
with fixed radial supports gives the spatial organization summary in
Table~\ref{tab:slicing_geometry}. The case/block matching analysis uses
$3\times20\times4=240$ comparisons, not 240 training runs.

Let $\boldsymbol a_i,\boldsymbol b_j\in\mathbb R^{4624}$ denote the 32
unlabeled head--slice supports from two representations of the same case and
block. With $\widetilde{\boldsymbol a}_i=\boldsymbol a_i/
\max(\|\boldsymbol a_i\|_2,10^{-15})$ and likewise for
$\widetilde{\boldsymbol b}_j$, the clipped cosine matrix is
$C_{ij}=\operatorname{clip}(\widetilde{\boldsymbol a}_i^{\mathsf T}
\widetilde{\boldsymbol b}_j,0,1)$. Hungarian matching selects the one-to-one
permutation $\pi^*$ that maximizes $\sum_{i=1}^{32}C_{i,\pi(i)}$, and
$S(\boldsymbol a,\boldsymbol b)=32^{-1}\sum_i C_{i,\pi^*(i)}$.
$S_{\mathrm{repeat}}$ compares learned supports from repeats 1 and 2 for the
same training size, development case, and block; $S_{\mathrm{fixed}}$ compares
learned and fixed-radial supports. Thus the 240 values comprise three training
sizes, 20 cases, and four blocks.
\begin{table}[!htbp]
\centering
\small
\caption{Geometry of learned and fixed radial slicing supports. Numbers of fitting cases $N=25,200,780$; two repeats; common development cases. Entries are ensemble medians.}
\label{tab:slicing_geometry}
\setlength{\tabcolsep}{5pt}
\begin{tabular}{@{}lrr@{}}
\toprule
Quantity & Learned slicing & Fixed radial slicing\\
\midrule
Radial support RMS (cm) & 2.591 & 0.292\\
Centroid pairwise distance (cm) & 1.113 & 3.445\\
Effective occupied slices & 3.747 & 7.913\\
\bottomrule
\end{tabular}
\end{table}

For a support with nonnegative weights $w_x$, define
\[
m=\max\left(\sum_x w_x,10^{-15}\right),\qquad
q_x=\frac{w_x}{m},\qquad
\boldsymbol\mu=\sum_xq_x\boldsymbol x .
\]
Radial support RMS is $[\sum_xq_x(r_x-\mu_r)^2]^{1/2}$. Centroid pairwise distance is the mean
Euclidean distance over the 28 unique pairs of the eight slice centroids within
each head. Effective occupied slices is
$\exp[-\sum_{s=1}^{8}p_s\log p_s]$, where $p_s$ is the active-token mean
assignment probability for slice $s$, averaged over the four heads. The table
aggregates the matched training sizes, repeats, development cases, blocks, and
supports used in Fig.~\ref{fig:transolver_slicing_final}.

\subsection{Reference-field variability and PCA projection}
Across $(n_e,n_i,\phi,T_e)$, the global normalized parameter-variability
ratios between the 2D and 1D reference ensembles are respectively
$(1.004,1.003,1.528,1.428)$. The reference-only PCA spectra in
Fig.~\ref{fig:solution_manifold_final} and projection comparison in
Fig.~\ref{fig:pca_rank96_matched} are evaluated on the corresponding
training and held-out cases. Table~\ref{tab:pca_rank96} reports the
saved-basis rank-96 projection/prediction comparison for 2D PCA-Net at
$N=800$, without refitting the stored normalized basis.
\begin{table}[!htbp]
\centering
\small
\caption{PCA-Net rank-96 errors using the saved basis. Projection uses the saved joint normalized basis and inverse transform; ratios are medians of case-wise ratios. Table~\ref{tab:main_performance_v8} first averages cases and the four fields within each repeat and then takes the five-repeat median, whereas this table reports per-field medians pooled over repeat--case observations. Consequently, averaging the four field medians here does not reproduce Table~\ref{tab:main_performance_v8}.}
\label{tab:pca_rank96}
\setlength{\tabcolsep}{5pt}
\begin{tabular}{@{}llrrr@{}}
\toprule
Field & Test set & Projection (\%) & Prediction (\%) & Ratio\\
\midrule
$n_e$ & interior & 0.09143 & 1.1586 & 13.34\\
$n_e$ & boundary & 0.23239 & 2.2033 & 10.11\\
$n_i$ & interior & 0.08196 & 1.0014 & 12.77\\
$n_i$ & boundary & 0.22337 & 2.0164 & 9.78\\
$\phi$ & interior & 0.01813 & 0.2500 & 13.53\\
$\phi$ & boundary & 0.04560 & 0.4798 & 11.12\\
$T_e$ & interior & 0.04908 & 0.5579 & 11.81\\
$T_e$ & boundary & 0.12446 & 1.2467 & 9.92\\
\bottomrule
\end{tabular}
\end{table}

The field-wise training-pool ranks accounting for 99\% of variance
are $(1,1,2,6)$ in 1D and $(2,2,4,7)$ in 2D for
$(n_e,n_i,\phi,T_e)$, respectively. The corresponding 99.9\% ranks are
$(2,2,3,10)$ and $(5,5,8,17)$. These measures describe the present
reference ensembles, not an isolated effect of changing spatial dimension.

\FloatBarrier


\begin{thebibliography}{99}

\bibitem{ref1}
Donnelly V M and Kornblit A 2013 Plasma etching: Yesterday, today, and tomorrow 
\textit{J. Vac. Sci. Technol. A} \textbf{31} 050825. 
doi: 10.1116/1.4819316.

\bibitem{ref2}
Oehrlein G S, Brandstadter S M, Bruce R L, Chang J P, DeMott J C, Donnelly V M, Dussart R, Fischer A, Gottscho R A, Hamaguchi S \textit{et al} 2024 Future of plasma etching for microelectronics: Challenges and opportunities 
\textit{J. Vac. Sci. Technol. B} \textbf{42} 041501. 
doi: 10.1116/6.0003579.

\bibitem{ref45}
Versolato O, Kaganovich I, Bera K, Lill T, Lee H-C, Hoekstra R, Sheil J and Nam S K 2024
Plasma sources for advanced semiconductor applications
\textit{Appl. Phys. Lett.} \textbf{125} 230401.
doi: 10.1063/5.0247819.

\bibitem{ref46}
Cardinaud C, Peignon M-C and Tessier P-Y 2000
Plasma etching: principles, mechanisms, application to micro- and nano-technologies
\textit{Appl. Surf. Sci.} \textbf{164} 72--83.
doi: 10.1016/S0169-4332(00)00328-7.

\bibitem{ref48}
Wang L, Wen D-Q, Tian C-B, Song Y-H and Wang Y-N 2021
Electron heating dynamics and plasma parameters control in capacitively coupled plasma
\textit{Acta Phys. Sin.} \textbf{70} 095214.
doi: 10.7498/aps.70.20210473.

\bibitem{ref5}
Rauf S, Bera K and Collins K 2010 Power dynamics in a low pressure capacitively coupled plasma discharge 
\textit{Plasma Sources Sci. Technol.} \textbf{19} 015014. 
doi: 10.1088/0963-0252/19/1/015014.

\bibitem{ref6}
Alves L L, Bogaerts A, Guerra V and Turner M M 2018 Foundations of modelling of nonequilibrium low-temperature plasmas 
\textit{Plasma Sources Sci. Technol.} \textbf{27} 023002. 
doi: 10.1088/1361-6595/aaa86d.

\bibitem{ref4}
Kim C H, Kim H, Park G and Lee H J 2021 The formation mechanism of nonuniformity from 2D nonlocal particle-dynamics in capacitive RF discharges 
\textit{Plasma Sources Sci. Technol.} \textbf{30} 065031. 
doi: 10.1088/1361-6595/ac0a47.

\bibitem{ref7}
Wang L, Hartmann P, Donkó Z, Song Y H and Schulze J 2025 Plasma uniformity control in capacitive RF discharges by an individually RF driven sidewall electrode 
\textit{Plasma Sources Sci. Technol.} \textbf{34} 045004. 
doi: 10.1088/1361-6595/adc504.

\bibitem{ref8}
Rauf S, Ganta S, Shi X, Bera K and Kenney J 2024 Ar/CF$_4$ capacitively coupled plasma generated using 40 MHz sinusoidal and 800 kHz rectangular waveform voltages 
\textit{Plasma Sources Sci. Technol.} \textbf{33} 125001. 
doi: 10.1088/1361-6595/ad95b9.

\bibitem{ref9}
Wang Y, Zhou Y, Chen J, Cao Y, Wang Z, Huang X and Zhang Y 2024 Numerical characterization of capacitively coupled plasma driven by tailored frequency modulated radio frequency source 
\textit{Plasma Sources Sci. Technol.} \textbf{33} 055009. 
doi: 10.1088/1361-6595/ad4587.

\bibitem{ref49}
Govindan T R and Meyyappan M 1995
One-Dimensional Modeling Studies of the Gaseous Electronics Conference RF Reference Cell
\textit{J. Res. Natl. Inst. Stand. Technol.} \textbf{100} 463--472.
doi: 10.6028/jres.100.035.

\bibitem{ref50}
Boeuf J P and Pitchford L C 1995
Two-dimensional model of a capacitively coupled rf discharge and comparisons with experiments in the Gaseous Electronics Conference reference reactor
\textit{Phys. Rev. E} \textbf{51} 1376--1390.
doi: 10.1103/PhysRevE.51.1376.

\bibitem{ref51}
Turner M M, Derzsi A, Donk\'o Z, Eremin D, Kelly S J, Lafleur T and Mussenbrock T 2013
Simulation benchmarks for low-pressure plasmas: capacitive discharges
\textit{Phys. Plasmas} \textbf{20} 013507.
doi: 10.1063/1.4775084.

\bibitem{ref52}
Ventzek P L G, Sommerer T J, Hoekstra R J and Kushner M J 1993
Two-dimensional hybrid model of inductively coupled plasma sources for etching
\textit{Appl. Phys. Lett.} \textbf{63} 605--607.
doi: 10.1063/1.109963.

\bibitem{ref53}
Wang H-Y, Jiang W and Wang Y-N 2010
Implicit and electrostatic Particle-in-cell/Monte Carlo model in two dimensional and axisymmetric geometry I: analysis of numerical techniques
\textit{Plasma Sources Sci. Technol.} \textbf{19} 045023.
doi: 10.1088/0963-0252/19/4/045023.

\bibitem{ref14}
Kushner M J 2009 Hybrid modelling of low temperature plasmas for fundamental investigations and equipment design 
\textit{J. Phys. D: Appl. Phys.} \textbf{42} 194013. 
doi: 10.1088/0022-3727/42/19/194013.

\bibitem{ref54}
Kushner M J \textit{et al} 1996
A three-dimensional model for inductively coupled plasma etching reactors: Azimuthal symmetry, coil properties, and comparison to experiments
\textit{J. Appl. Phys.} \textbf{80} 1337--1344.
doi: 10.1063/1.362932.

\bibitem{ref55}
Hoekstra R J, Grapperhaus M J and Kushner M J 1997
Integrated plasma equipment model for polysilicon etch profiles in an inductively coupled plasma reactor with subwafer and superwafer topography
\textit{J. Vac. Sci. Technol. A} \textbf{15} 1913--1921.
doi: 10.1116/1.580659.

\bibitem{ref56}
Zhang D and Kushner M J 2000
Surface kinetics and plasma equipment model for Si etching by fluorocarbon plasmas
\textit{J. Appl. Phys.} \textbf{87} 1060--1069.
doi: 10.1063/1.371980.

\bibitem{ref57}
Lymberopoulos D P and Economou D J 1995
Two-Dimensional Self-Consistent Radio Frequency Plasma Simulations Relevant to the Gaseous Electronics Conference RF Reference Cell
\textit{J. Res. Natl. Inst. Stand. Technol.} \textbf{100} 473--494.
doi: 10.6028/jres.100.036.

\bibitem{ref58}
Chiu Y-M, Chiang C-H, Hung C-T, Hu M-H, Wu J-S and Hwang F-N 2014
Parallel 2D Axisymmetric Fluid Modeling of CF$_4$ Discharge in an Inductively Coupled Plasma Source During SiO$_2$ Etching
\textit{Plasma Process. Polym.} \textbf{11} 366--390.
doi: 10.1002/ppap.201300134.

\bibitem{ref59}
Chen K-L, Tseng M-F, Lo M-C, Hamaguchi S, Hu M-H, Lee Y-M and Wu J-S 2021
Development of a Massively Parallelized Fluid-Based Plasma Simulation Code With a Finite-Volume Method on an Unstructured Grid
\textit{IEEE Trans. Plasma Sci.} \textbf{49} 104--119.
doi: 10.1109/TPS.2020.3013632.

\bibitem{ref60}
Turkoz E and Celik M 2015
AETHER: A simulation platform for inductively coupled plasma
\textit{J. Comput. Phys.} \textbf{286} 87--102.
doi: 10.1016/j.jcp.2015.01.027.

\bibitem{ref61}
Wang H-Y, Jiang W and Wang Y-N 2009
Parallelization and optimization of electrostatic Particle-in-Cell/Monte-Carlo Coupled codes as applied to RF discharges
\textit{Comput. Phys. Commun.} \textbf{180} 1305--1314.
doi: 10.1016/j.cpc.2009.02.009.

\bibitem{ref62}
Shin J H, Kim H and Lee H J 2022
Two-dimensional particle-in-cell simulation parallelized with graphics processing units for the investigation of plasma kinetics in a dual-frequency capacitively coupled plasma
\textit{Rev. Mod. Plasma Phys.} \textbf{6} 30.
doi: 10.1007/s41614-022-00092-2.

\bibitem{ref63}
Li J-Z, Zhao M-L, Zhang Y-R, Gao F and Wang Y-N 2025
Fast simulation strategy for capacitively-coupled plasmas based on fluid model
\textit{Comput. Phys. Commun.} \textbf{307} 109392.
doi: 10.1016/j.cpc.2024.109392.

\bibitem{refTsagkaridis2025}
Tsagkaridis M, Oliver T A, Fries D, O'Connor R,
Barberena-Valencia J P, Raja L L, Varghese P L
and Moser R D 2025
Modeling of low-temperature argon plasma in
capacitively-coupled glow discharges with a
collisional-radiative model
\textit{Plasma Sources Sci. Technol.}
\textbf{34} 105013.
doi: 10.1088/1361-6595/ae0c33.

\bibitem{ref35}
Kanarik K J, Osowiecki W T, Lu Y, Talukder D, Roschewsky N, Park S N,
Kamon M, Fried D M and Gottscho R A 2023
Human--machine collaboration for improving semiconductor process development
\textit{Nature} \textbf{616} 707--711.
doi: 10.1038/s41586-023-05773-7.

\bibitem{ref40}
Ko J, Bae J, Park M, Jo Y, Lee H, Kim K, Yoo S, Nam S K, Sung D and Kim B 2023
Computational approach for plasma process optimization combined with deep learning model
\textit{J. Phys. D: Appl. Phys.} \textbf{56} 344001.
doi: 10.1088/1361-6463/acd1fd.

\bibitem{ref28}
Myung S \textit{et al} 2026 Simulation-guided AI-driven digital twin for plasma etching in semiconductor fabrication 
\textit{Device} 
\textbf{4} 101026. 
doi: 10.1016/j.device.2025.101026.

\bibitem{ref66}
Conlin P, Sridhar S and Ventzek P 2026
Designing pulsed plasma etch processes by genetic algorithm
\textit{J. Vac. Sci. Technol. A} \textbf{44} 033001.
doi: 10.1116/6.0005164.

\bibitem{ref64}
Osowiecki W T \textit{et al} 2024
Achieving Sustainability in the Semiconductor Industry: The Impact of Simulation and AI
\textit{IEEE Trans. Semicond. Manuf.} \textbf{37} 464--474.
doi: 10.1109/TSM.2024.3438622.

\bibitem{ref17}
Bonzanini A D, Shao K, Graves D B, Hamaguchi S and Mesbah A 2023 Foundations of machine learning for low-temperature plasmas: methods and case studies 
\textit{Plasma Sources Sci. Technol.} \textbf{32} 024003. 
doi: 10.1088/1361-6595/acb28c.

\bibitem{ref25}
Anirudh R \textit{et al} 2023 2022 Review of Data-Driven Plasma Science
\textit{IEEE Trans. Plasma Sci.} 
\textbf{51} 1750--1838. 
doi: 10.1109/TPS.2023.3268170.

\bibitem{ref29}
Trieschmann J, Vialetto L and Gergs T 2023 
Machine learning for advancing low-temperature plasma modeling and simulation 
\textit{J. Micro/Nanopattern. Mater. Metrol.} 
\textbf{22} 041504.
doi: 10.1117/1.JMM.22.4.041504.

\bibitem{ref31}
Gopakumar V \textit{et al} 2024 Plasma surrogate modelling using Fourier neural operators 
\textit{Nuclear Fusion} 
\textbf{64} 056025. 
doi: 10.1088/1741-4326/ad313a.

\bibitem{ref27}
Liu P, Wu Q, Ren X, Wang Y and Ni D 2024 A deep-learning-based surrogate modeling method with application to plasma processing 
\textit{Chemical Engineering Research and Design} 
\textbf{211} 299--317. 
doi: 10.1016/j.cherd.2024.09.031.

\bibitem{ref18}
Li J, Jiang X, He Q, Zhang Y, Wu T, Jiang S, Song Y and Jia W 2025 Capacitively coupled argon plasmas fluid simulations with deep learning surrogate model: Asymmetric inference and quantitative trust boundaries 
\textit{Acta Phys. Sin.} \textbf{74} 235205. 
doi: 10.7498/aps.74.20251290.

\bibitem{ref67}
Dasbach S and Wiesen S 2023
Towards fast surrogate models for interpolation of tokamak edge plasmas
\textit{Nucl. Mater. Energy} \textbf{34} 101396.
doi: 10.1016/j.nme.2023.101396.

\bibitem{ref33}
Carey N \textit{et al} 2025 Neural operator surrogate models of plasma edge simulations: feasibility and data efficiency 
\textit{Nuclear Fusion} 
\textbf{65} 106010. 
doi: 10.1088/1741-4326/adfdfb.

\bibitem{ref32}
Diaw A \textit{et al} 2024 Efficient learning of accurate surrogates for simulations of complex systems 
\textit{Nature Machine Intelligence} 
\textbf{6} 568--577. 
doi: 10.1038/s42256-024-00839-1.

\bibitem{ref41}
de Hoop M V, Huang D Z, Qian E and Stuart A M 2022
The Cost-Accuracy Trade-Off in Operator Learning with Neural Networks
\textit{Journal of Machine Learning} \textbf{1} 299--341.
doi: 10.4208/jml.220509.

\bibitem{refCost}
Huang D Z and Stuart A M 2026
Cost-Accuracy Trade-offs: Neural Operator vs Classical Numerical Solver 
\textit{arXiv preprint}.  
arXiv:2609.24021.

\bibitem{ref68}
Pestourie R, Mroueh Y, Nguyen T V, Das P and Johnson S G 2020
Active learning of deep surrogates for PDEs: application to metasurface design
\textit{npj Comput. Mater.} \textbf{6} 164.
doi: 10.1038/s41524-020-00431-2.

\bibitem{ref69}
Wang S, Wang H and Perdikaris P 2021
Learning the solution operator of parametric partial differential equations with physics-informed DeepONets
\textit{Sci. Adv.} \textbf{7} eabi8605.
doi: 10.1126/sciadv.abi8605.

\bibitem{ref70}
Pestourie R, Mroueh Y, Rackauckas C, Das P and Johnson S G 2023
Physics-enhanced deep surrogates for partial differential equations
\textit{Nat. Mach. Intell.} \textbf{5} 1458--1465.
doi: 10.1038/s42256-023-00761-y.

\bibitem{ref71}
Lu L, Pestourie R, Johnson S G and Romano G 2022
Multifidelity deep neural operators for efficient learning of partial differential equations with application to fast inverse design of nanoscale heat transport
\textit{Phys. Rev. Research} \textbf{4} 023210.
doi: 10.1103/PhysRevResearch.4.023210.

\bibitem{ref72}
Chen W, Song J, Ren P, Subramanian S, Morozov D and Mahoney M W 2024
Data-Efficient Operator Learning via Unsupervised Pretraining and In-Context Learning
\textit{Adv. Neural Inf. Process. Syst.} \textbf{37} 6213--6245.
doi: 10.52202/079017-0201.

\bibitem{ref15}
Liang Y S, Liu Y X, Zhang Y R and Wang Y N 2015 Fluid simulation and experimental validation of plasma radial uniformity in 60 MHz capacitively coupled nitrogen discharges 
\textit{J. Appl. Phys.} \textbf{117} 083301. 
doi: 10.1063/1.4913221.

\bibitem{ref75}
Liu Y-X, Zhang Y-R, Bogaerts A and Wang Y-N 2015
Electromagnetic effects in high-frequency large-area capacitive discharges: A review
\textit{J. Vac. Sci. Technol. A} \textbf{33} 020801.
doi: 10.1116/1.4907926.

\bibitem{ref76}
Zhang Y-R, Xu X, Bogaerts A and Wang Y-N 2012
Fluid simulation of the phase-shift effect in hydrogen capacitively coupled plasmas: II. Radial uniformity of the plasma characteristics
\textit{J. Phys. D: Appl. Phys.} \textbf{45} 015203.
doi: 10.1088/0022-3727/45/1/015203.

\bibitem{ref38}
Sitzmann V, Martel J N P, Bergman A W, Lindell D B and Wetzstein G 2020
Implicit neural representations with periodic activation functions
\textit{Advances in Neural Information Processing Systems} \textbf{33} 7462--7473.

\bibitem{ref36}
Berkooz G, Holmes P and Lumley J L 1993
The proper orthogonal decomposition in the analysis of turbulent flows
\textit{Annu. Rev. Fluid Mech.} \textbf{25} 539--575.
doi: 10.1146/annurev.fl.25.010193.002543.

\bibitem{ref19}
Hesthaven J S and Ubbiali S 2018 Non-intrusive reduced order modeling of nonlinear problems using neural networks 
\textit{J. Comput. Phys.} \textbf{363} 55--78. 
doi: 10.1016/j.jcp.2018.02.037.

\bibitem{bhattacharya2021model}
Bhattacharya K, Hosseini B, Kovachki N B and Stuart A M 2021
Model reduction and neural networks for parametric PDEs
\textit{SMAI J. Comput. Math.} \textbf{7} 121--157.
doi: 10.5802/smai-jcm.74.

\bibitem{ref23}
Ronneberger O, Fischer P and Brox T 2015 U-Net: Convolutional networks for biomedical image segmentation 
\textit{MICCAI 2015, Lecture Notes in Computer Science} 
\textbf{9351} 234--241. 
doi: 10.1007/978-3-319-24574-4\_28.

\bibitem{ref22}
Li Z, Kovachki N, Azizzadenesheli F, Liu B, Bhattacharya K, Stuart A and Anandkumar A 2021 Fourier neural operator for parametric partial differential equations 
\textit{International Conference on Learning Representations (ICLR 2021)}.

\bibitem{ref21}
Kovachki N, Li Z, Liu B, Azizzadenesheli F, Bhattacharya K, Stuart A and Anandkumar A 2023 Neural operator: Learning maps between function spaces with applications to PDEs 
\textit{J. Mach. Learn. Res.} \textbf{24} 1--97.

\bibitem{ref24}
Wu H, Luo H, Wang H, Wang J and Long M 2024 Transolver: A fast Transformer solver for PDEs on general geometries 
\textit{Proc. 41st Int. Conf. on Machine Learning} 
\textbf{235} 53681--53705.

\bibitem{ref74}
Chen H, Wu R, Grinspun E, Zheng C and Chen P Y 2023
Implicit Neural Spatial Representations for Time-dependent PDEs
\textit{Proc. 40th Int. Conf. on Machine Learning} \textbf{202} 5162--5177.

\bibitem{ref20}
Fresca S and Manzoni A 2022 POD-DL-ROM: Enhancing deep learning-based reduced order models for nonlinear parametrized PDEs by proper orthogonal decomposition 
\textit{Comput. Methods Appl. Mech. Eng.} \textbf{388} 114181. 
doi: 10.1016/j.cma.2021.114181.

\bibitem{ref42}
Peng Z, Wang M and Li F 2023
A learning-based projection method for model order reduction of transport problems
\textit{J. Comput. Appl. Math.} \textbf{418} 114560.
doi: 10.1016/j.cam.2022.114560.

\bibitem{ref37}
Guo X, Li W and Iorio F 2016
Convolutional neural networks for steady flow approximation
\textit{Proc. 22nd ACM SIGKDD Int. Conf. Knowledge Discovery and Data Mining}
481--490.
doi: 10.1145/2939672.2939738.

\bibitem{ref43}
Fan Y, Lin L, Ying L and Zepeda-N\'{u}\~{n}ez L 2019
A multiscale neural network based on hierarchical matrices
\textit{Multiscale Model. Simul.} \textbf{17} 1189--1213.
doi: 10.1137/18M1203602.

\bibitem{ref44}
Fan Y, Orozco Bohorquez C and Ying L 2019
BCR-Net: A neural network based on the nonstandard wavelet form
\textit{J. Comput. Phys.} \textbf{384} 1--15.
doi: 10.1016/j.jcp.2019.02.002.


\bibitem{li2023geofno}
Li Z, Huang D Z, Liu B and Anandkumar A 2023
Fourier neural operator with learned deformations for PDEs on general geometries
\textit{J. Mach. Learn. Res.} \textbf{24}(388) 1--26.

\bibitem{hu2026transolverlinear}
Hu W, Liu S, Qiao P, Sun Z and Dou Y 2026
Transolver is a linear Transformer: Revisiting Physics-Attention
through the lens of linear attention
\textit{Proc. AAAI Conf. Artif. Intell.} \textbf{40}(1) 408--416.
doi: 10.1609/aaai.v40i1.37003.

\bibitem{lanthaler2023pcanet}
Lanthaler S 2023
Operator learning with PCA-Net: upper and lower complexity bounds
\textit{J. Mach. Learn. Res.} \textbf{24}(318) 1--67.

\bibitem{sakiyama2006corona}
Sakiyama Y and Graves D B 2006 Corona-glow transition in the atmospheric
pressure RF-excited plasma needle \textit{J. Phys. D: Appl. Phys.}
\textbf{39} 3644--3652.
doi: 10.1088/0022-3727/39/16/018.

\bibitem{dechant2023zapdos}
DeChant C, Icenhour C, Keniley S, Gall G, Lindsay A, Curreli D and Shannon S
2023 Verification and validation of the open-source plasma fluid code:
Zapdos \textit{Comput. Phys. Commun.} \textbf{291} 108837.
doi: 10.1016/j.cpc.2023.108837.

\bibitem{lymberopoulos1993_2d}
Lymberopoulos D P and Economou D J 1993 Fluid simulations of radio frequency
glow discharges: Two-dimensional argon discharge including metastables
\textit{Appl. Phys. Lett.} \textbf{63} 2478--2480.
doi: 10.1063/1.110454.

\bibitem{lymberopoulos1994_modeling}
Lymberopoulos D P and Economou D J 1994 Modeling and simulation of glow
discharge plasma reactors \textit{J. Vac. Sci. Technol. A}
\textbf{12} 1229--1236.
doi: 10.1116/1.579300.


\end{thebibliography}
\end{document}